\documentclass[journal,twoside,web]{ieeecolor}
\usepackage{generic}
\usepackage{cite}
\usepackage{amsmath,amssymb,amsfonts}
\usepackage{epstopdf}
\usepackage[pdftex,final]{graphicx}
\usepackage{graphicx}
\usepackage{algorithm,algpseudocode}
\usepackage{hyperref}
\hypersetup{hidelinks=true}
\usepackage{textcomp}
\usepackage{amsmath}
\usepackage{algorithm}
\usepackage{array}
\usepackage{multirow}

\def\BibTeX{{\rm B\kern-.05em{\sc i\kern-.025em b}\kern-.08em
    T\kern-.1667em\lower.7ex\hbox{E}\kern-.125emX}}
\begin{document}
\title{DCS-ST for Classification of Breast Cancer
Histopathology Images with Limited Annotations}
\author{LIU SUXING\IEEEmembership{},
        BYUNGWON MIN\IEEEmembership{}
\thanks{LIU SUXING is with the School of Digital Arts, Jiangxi Arts \& Ceramics Technology Institute, Jindezhen 33001, China; and also with the Department of IT Engineering, Mokwon University, Daejeon 35349, South Korea (e-mail: bentondoucet@gmail.com).}
\thanks{BYUNGWON MIN is with the Department of IT Engineering, Mokwon University, Daejeon 35349, South Korea}}

\maketitle

\begin{abstract}
Deep learning methods have shown promise in classifying breast cancer histopathology images, but their performance often declines with limited annotated data, a critical challenge in medical imaging due to the high cost and expertise required for annotations. To address this, we propose the Dynamic Cross-Scale Swin Transformer (DCS-ST), a robust approach designed for classifying breast cancer histopathology images under constrained annotation settings. DCS-ST processes input images by leveraging a small set of labeled data alongside a larger pool of unlabeled data, employing a pseudo-labeling strategy to generate high-confidence labels for training. A dynamic window predictor adaptively adjusts attention window sizes across scales, enhancing the Swin Transformer backbone’s ability to capture both local and global features. Additionally, a cross-scale attention module fuses multi-scale features via multi-head attention, boosting representational capacity. A semi-supervised learning strategy, incorporating forward diffusion and reverse denoising, further stabilizes feature extraction and reduces inference uncertainty. Extensive experiments on the BreakHis, ICIAR2018 BACH Challenge, and Mini-DDSM datasets demonstrate that DCS-ST consistently outperforms state-of-the-art methods across various magnifications and classification tasks.
\end{abstract}

\begin{IEEEkeywords}
Breast Cancer Classification, Cross-Scale Attention, Dynamic Window Prediction, Histopathology Images, Medical Image Analysis, Pseudo-Labeling, Semi-Supervised Learning, Swin Transformer.
\end{IEEEkeywords}

\section{Introduction}
\label{sec:introduction}

Histopathology images of breast cancer, often captured under weakly labeled conditions. These present significant analytical challenges that impair the performance of downstream vision tasks such as automated classification, lesion detection, and clinical diagnostic assistance\cite{b1}. As one of the most prevalent and life-threatening cancers worldwide, breast cancer demands early and accurate diagnosis to improve patient outcomes, with histopathology images serving as the gold standard due to their rich cellular and tissue detail. However, annotating these images relies heavily on pathologists’ expertise and time, resulting in a critical scarcity of labeled data. This data insufficiency hampers the robustness of conventional deep learning models, which typically require vast annotated datasets for effective training\cite{b2}.

In recent years, Transformer-based models like the Swin Transformer\cite{b25} have shown remarkable success in high-resolution image tasks, leveraging their hierarchical structure and shifted window mechanism to efficiently capture local and global features, making them highly suitable for analyzing complex histopathology images. Yet, in scenarios with limited labeled data, optimizing these models to handle diverse tissue structures and pathological patterns remains a formidable challenge. Traditional approaches often depend on hand-crafted feature extraction rules, which struggle to adapt to the complexity and variability of histopathology images\cite{b3}.

To address this, we propose the Dynamic Cross-Scale Swin Transformer\cite{b25} (DCS-ST), a robust approach tailored for classifying breast cancer histopathology images under weakly labeled conditions. Building on the Swin Transformer architecture, DCS-ST leverages a small set of labeled data alongside a larger pool of unlabeled data, employing a semi-supervised learning strategy with pseudo-labeling to generate high-confidence labels for training. Our method introduces two novel components: a Dynamic Window Predictor and a Cross-Scale Attention Module. The Dynamic Window Predictor uses a convolutional layer to adaptively predict window sizes for each image region, enabling the model to dynamically adjust its attention mechanism based on content. The Cross-Scale Attention Module integrates multi-scale features via a multi-head attention mechanism, significantly enhancing the model’s ability to discern intricate patterns, such as normal, benign, in situ carcinoma, and invasive carcinoma, across varying spatial resolutions. Compared to the standard Swin Transformer, DCS-ST exhibits superior feature representation and classification stability when labeled data is scarce.

We evaluate DCS-ST on multiple benchmark datasets, including BreakHis\cite{b6}, Mini-DDSM\cite{b7}, and  ICIAR2018 BACH Challenge\cite{b8}, covering two-class, three-class, and four-class breast cancer classification tasks. For the ICIAR2018 dataset, which includes four classes (normal, benign, in situ carcinoma, and invasive carcinoma), we simulate real-world weakly labeled scenarios by partitioning the dataset into 80\% for training (with only 5\% labeled and 95\% unlabeled) and 20\% for testing, using stratified sampling to ensure class balance. Experimental results demonstrate that DCS-ST consistently outperforms existing methods across key metrics, including AUC-ROC\cite{b9}, balanced accuracy\cite{b10}, F1 score\cite{b11}, and Cohen’s Kappa\cite{b12}, delivering high-quality classification outcomes even with minimal labeled data, while avoiding overfitting issues common in supervised methods and instability seen in unsupervised ones.

Our contributions are summarized as follows:

\begin{itemize}
\item We propose the Dynamic Cross-Scale Swin Transformer (DCS-ST), which leverages dynamic window prediction, cross-scale attention, and semi-supervised learning for robust histopathology image classification under limited annotations.
\item We design innovative Dynamic Window Predictor and Cross-Scale Attention Modules to enhance the model’s ability to capture complex tissue features across scales.
\item Extensive experiments on the BreakHis,  Mini-DDSM datasets, and ICIAR2018 BACH Challenge validate the method’s superior performance across various classification tasks and magnifications.
\end{itemize}
\section{Related Works}

\subsection{Semi-Supervised Learning for Image Classification}

Semi-supervised learning (SSL) has become a vital approach for image classification, particularly in scenarios with limited labeled data, such as medical imaging, where high-quality annotations are costly and time-intensive to acquire. SSL leverages a small labeled dataset alongside a larger pool of unlabeled data to improve model performance. Traditional SSL methods include self-training~\cite{b13,b14}, which generates pseudo-labels for unlabeled data using the model’s predictions and incorporates them into training, co-training~\cite{b15}, which uses complementary views or models to refine pseudo-labels, and graph-based approaches, which propagate labels through similarity graphs between samples.

With the rise of deep learning, SSL methods have evolved significantly. FixMatch~\cite{b16} combines pseudo-labeling with consistency regularization, enforcing consistent predictions across augmented versions of unlabeled data, achieving strong performance in image classification. MixMatch~\cite{b17} integrates data augmentation and MixUp to enhance model generalization by blending labeled and unlabeled samples. Unsupervised Data Augmentation (UDA)~\cite{b18} further emphasizes augmentation invariance, using diverse transformations to improve robustness in SSL settings.

SSL has been increasingly applied to address data scarcity in medical image classification. For breast cancer histopathology image classification, Abdulrazzaq et al.
\cite{b19} proposed an SSL framework that leverages unlabeled data to enhance model robustness, achieving improved performance on datasets with limited annotations. Pani et al.\cite{b20} combined SSL with transfer learning, applying it to medical image classification tasks with constrained labeled data, demonstrating its effectiveness in handling diverse medical imaging modalities. These works highlight the potential of SSL in medical imaging. Still, they often rely on generic architectures, leaving room for specialized models tailored to histopathology images, as addressed in our work with the Dynamic Cross-Scale Swin Transformer (DCS-ST), which integrates pseudo-labeling and semi-supervised learning to tackle limited annotations in breast cancer classification.

\begin{figure*}
    \centering
    \includegraphics[width=0.4\textwidth,keepaspectratio]{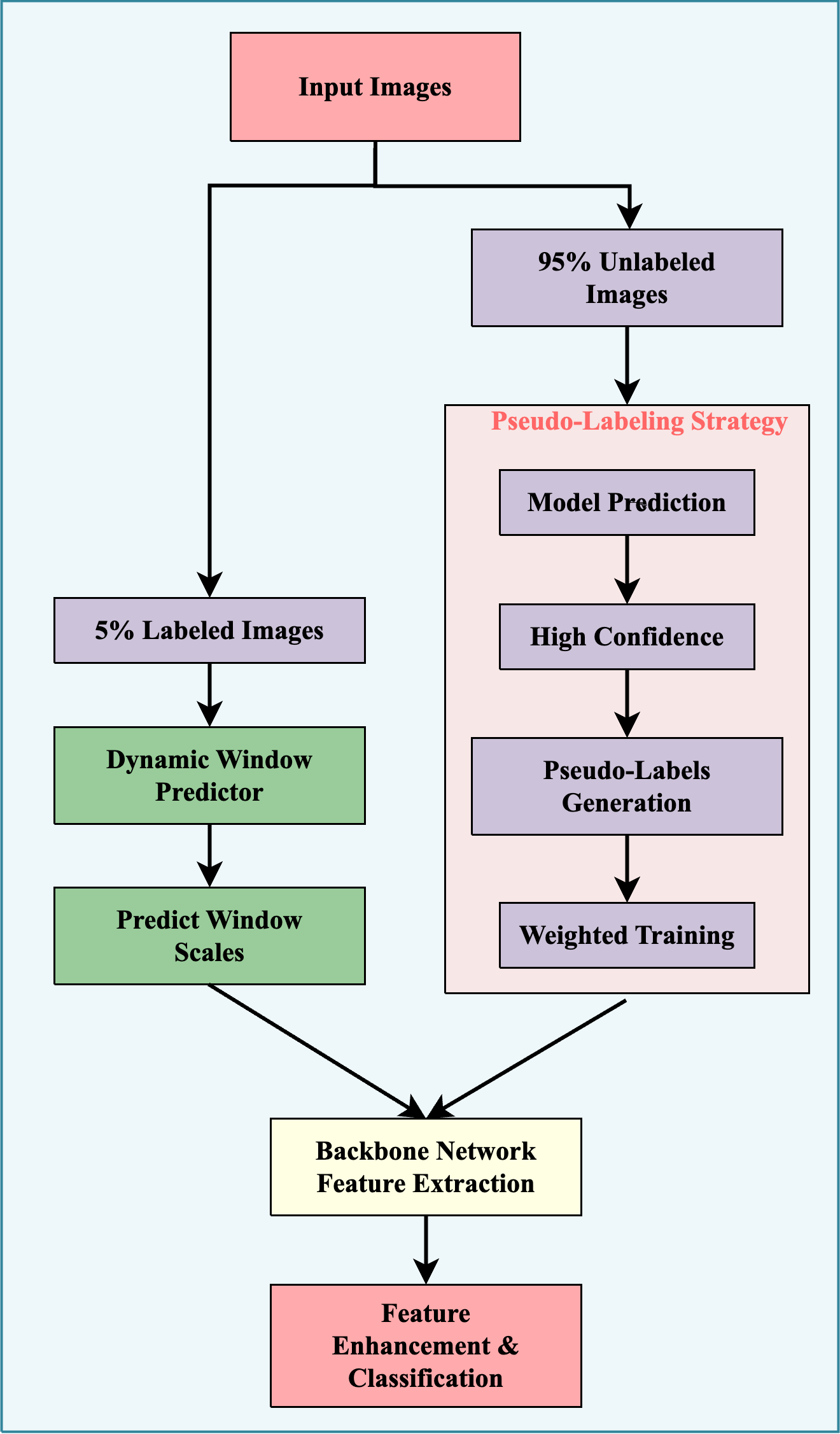} 

    \caption{Flowchart of the Dynamic Cross-Scale Swin Transformer (DCS-ST) pipeline for breast cancer histopathology image classification with limited annotations. Input images are split into 5\% labeled and 95\% unlabeled data, processed via a Dynamic Window Predictor and pseudo-labeling strategy, followed by backbone feature extraction and classification. This pipeline effectively leverages both labeled and unlabeled data for robust performance.}
    \label{figc}
\end{figure*}

\subsection{Transformers in Medical Image Analysis}

Transformer models, originally developed for natural language processing, have been adapted for computer vision with significant success. The Vision Transformer (ViT)\cite{b24} pioneered this transition by segmenting images into patches and treating them as sequences, achieving competitive performance on benchmark datasets like ImageNet. However, ViT’s global self-attention mechanism can be computationally intensive for high-resolution images, such as those in medical imaging.

The Swin Transformer\cite{b25}, a hierarchical Transformer with a shifted window mechanism, addresses this limitation by capturing both local and global features efficiently, making it well-suited for high-resolution medical images. In medical image analysis, Swin Transformer has been applied to tasks like segmentation. Cao et al.\cite{b41} achieves state-of-the-art results by leveraging its hierarchical structure to model spatial relationships in medical images.

For breast cancer histopathology image classification, Transformer-based models are gaining attention. Nayak et al.\cite{b42} proposed a Transformer-based model for breast cancer image classification, using self-attention to capture critical pathological features, showing promising results on datasets like BreakHis. However, these methods often assume access to large labeled datasets, which is not feasible in many real-world medical scenarios. Our work builds on the Swin Transformer by introducing dynamic window prediction and cross-scale attention, specifically designed to handle limited annotations in breast cancer histopathology image classification across datasets like BreakHis, ICIAR2018 BACH Challenge, and Mini-DDSM.

\subsection{Dynamic Window and Cross-Scale Attention Mechanisms}

Dynamic window and cross-scale attention mechanisms are emerging techniques in computer vision, aimed at improving the capture of multi-scale features. Dynamic window mechanisms enable models to adjust window sizes based on input content adaptively, providing flexibility in focusing on regions of interest at varying scales. Cross-scale attention mechanisms enhance feature representation by facilitating interactions between features at different scales, improving the model’s ability to handle complex patterns.

In general image classification, Yang et al.\cite{b43} proposed a Dynamic Window Attention Network that dynamically adjusts window sizes to improve classification accuracy, demonstrating its effectiveness in capturing context-aware features. Chen et al.\cite{b44} developed a cross-scale attention module to enhance information exchange between feature maps at different scales, boosting representational power for tasks like object recognition.

In medical image analysis, these mechanisms are still being explored but show significant potential. Shen et al.\cite{b45} applied dynamic window mechanisms to medical image segmentation, achieving improved precision by adaptively adjusting window sizes to focus on relevant anatomical structures. Zhao et al.\cite{b46} proposed a cross-scale attention network for medical image classification and segmentation, demonstrating its ability to integrate multi-scale features for better performance. While these methods highlight the benefits of dynamic and cross-scale attention, their application to breast cancer histopathology image classification under limited annotations remains underexplored. Our Dynamic Cross-Scale Swin Transformer (DCS-ST) addresses this gap by combining a Dynamic Window Predictor and a Cross-Scale Attention Module, enabling robust feature extraction and classification of histopathology images with minimal labeled data.

\begin{algorithm}
\caption{Enhanced Swin Transformer with Semi-Supervised Learning}
\begin{algorithmic}[0]
\Require Labeled dataset $D_{labeled} = \{(x_i, y_i)\}_{i=1}^{N_l}$, Unlabeled dataset $D_{unlabeled} = \{x_{u_j}\}_{j=1}^{N_u}$, Test dataset $D_{test}$, Epochs $E$, Threshold $\tau = 0.9$
\State Initialize Enhanced Swin Transformer parameters $\theta$, optimizer and scheduler

\For{epoch $e = 1, \ldots, E$}
    \If{$e < 2$} \Comment{Initial training on labeled data only}
        \State Train model on $D_{labeled}$, update parameters $\theta$
    \Else \Comment{Semi-supervised learning phase}
        \State Initialize PseudoLabels $\leftarrow \emptyset$, PseudoImages $\leftarrow \emptyset$
        
        \State \textbf{Generate pseudo-labels:} For all $x_u \in D_{unlabeled}$, predict classes and add samples with confidence $> \tau$ to pseudo-label set
        
        \State \textbf{Training:} Train model on $D_{labeled}$
        
        \If{PseudoLabels exist}
            \State Train model on pseudo-labeled data using weighted loss (0.8)
        \EndIf
    \EndIf
    \State Update learning rate scheduler
\EndFor

\State \textbf{Testing:} Compute predictions on $D_{test}$ and evaluate metrics

\State \Return Classification metrics and trained model
\end{algorithmic}
\end{algorithm}

\section{Method}

\subsection{Overview of the Method}
To tackle the challenge of classifying breast cancer histopathology images with limited annotated data, we introduce the Dynamic Cross-Scale Swin Transformer (DCS-ST). This novel architecture enhances the Swin Transformer framework. DCS-ST incorporates three key innovations: a dynamic window predictor, a cross-scale attention module, and a semi-supervised learning strategy. These components work synergistically to enable robust feature extraction and classification, effectively capturing local and global image features despite scarce annotations. The dynamic window predictor adapts the model's focus to varying feature scales, the cross-scale attention module integrates multi-scale information for richer representations, and the semi-supervised strategy leverages abundant unlabeled data to improve generalization. This pipeline, illustrated in Fig.1, addresses the critical need for accurate classification in data-constrained medical imaging scenarios.

\subsection{Dynamic Window Predictor}
The dynamic window predictor enhances the Swin Transformer's flexibility by dynamically adjusting attention window sizes based on the spatial content of the input image. Unlike the fixed window sizes in traditional Swin Transformers, this module enables the model to focus adaptively across different scales—a critical capability for histopathology images, where features range from small cellular structures to larger tissue patterns.

The predictor employs a lightweight convolutional layer to estimate window scales:
\[
\text{scales} = \text{softmax}(\text{Conv}_{1 \times 1}(x)),
\]
where \( x \in \mathbb{R}^{B \times C \times H \times W} \) is the input image, \( \text{Conv}_{1 \times 1} \) is a 1x1 convolution with output channels equal to the number of scales (e.g., 3), and \( \text{softmax} \) normalizes the predictions into a probability distribution. The output, \( \text{scales} \in \mathbb{R}^{B \times S \times H \times W} \), informs the attention mechanism, allowing scale-adaptive feature extraction. As depicted in Fig. 1, this process begins with the labeled data (5\% of the dataset), producing predicted scales that guide subsequent feature processing.

\subsection{Cross-Scale Attention Module}
The cross-scale attention module strengthens the model's representational power by fusing features from different transformer stages using a multi-head attention mechanism. This integration captures dependencies across various resolutions, enabling the model to identify subtle histopathological patterns that may signify different cancer subtypes, combining local details with global context efficiently.

For features from the current stage \( \text{current\_features} \in \mathbb{R}^{B \times C \times H \times W} \) and the previous stage \( \text{prev\_features} \in \mathbb{R}^{B \times C \times H \times W} \), the module processes them as follows:
\[
\text{current\_flat} = \text{reshape}(\text{current\_features}, [H \cdot W, B, C]),
\]
\[
\text{prev\_flat} = \text{reshape}(\text{prev\_features}, [H \cdot W, B, C]),
\]
\[
\text{attended} = \text{MultiheadAttention}(\text{current\_flat}, \text{prev\_flat}, \text{prev\_flat}),
\]
where the output \( \text{attended} \in \mathbb{R}^{H \cdot W \times B \times C} \) is reshaped back to \( \mathbb{R}^{B \times C \times H \times W} \). This mechanism ensures computationally efficient modeling of cross-scale relationships, enhancing feature extraction within the backbone network (see Fig. 1).

\subsection{Semi-Supervised Learning Strategy}
To address the limited availability of labeled data, we implement a semi-supervised learning strategy that leverages both labeled (5\%) and unlabeled (95\%) samples, as shown in Fig. 1. This approach employs a diffusion process to introduce controlled noise into unlabeled data, followed by denoising to stabilize feature extraction, and integrates pseudo-labeling to enhance training, as outlined in Algorithm 1.

The forward diffusion process adds noise incrementally:
\[
q(x_{1:T} | x_0) = \prod_{t=1}^T q(x_t | x_{t-1}),
\]
\[
q(x_t | x_{t-1}) = \mathcal{N}(x_t; \sqrt{1 - \beta_t} x_{t-1}, \beta_t I),
\]
where \( x_0 \) is the original unlabeled image, \( T \) is the number of diffusion steps, and \( \beta_t \) controls the noise schedule. The reverse denoising process reconstructs the data:
\[
p_\theta(x_{0:T}) = p(x_T) \prod_{t=1}^T p_\theta(x_{t-1} | x_t),
\]
where \( p_\theta \) is learned by the model. Additionally, pseudo-labels are generated for unlabeled images with confidence scores above a threshold (\( \tau = 0.9 \)), as per Algorithm 1, and incorporated into training with a weighted loss (factor 0.8). This dual approach enhances robustness and generalization by exploiting the unlabeled majority.

\subsection{Model Architecture}
The DCS-ST architecture integrates the Swin Transformer backbone with the proposed enhancements, forming a cohesive pipeline (see Fig. 1):
\begin{itemize}
    \item Swin Transformer Backbone: A pre-trained Swin-Base model extracts hierarchical features from the input image.
    \item Dynamic Window Predictor: Adjusts attention window sizes dynamically, guiding the backbone's attention mechanism.
    \item Cross-Scale Attention Module: Fuses multi-scale features across transformer stages for enhanced representation.
    \item Classifier: A linear layer maps the backbone's 1024-dimensional feature vector to four class logits, corresponding to breast cancer histopathology categories.
\end{itemize}
The input image is processed sequentially: the dynamic window predictor informs the backbone, the cross-scale attention module refines features, and the classifier delivers the final prediction, leveraging both labeled and pseudo-labeled data.

\subsection{Training and Evaluation}
The model is trained using the cross-entropy loss:
\[
L = -\frac{1}{N} \sum_{i=1}^N \sum_{c=1}^C y_{i,c} \log(p_{i,c}),
\]
where \( N \) is the batch size, \( C = 4 \) is the number of classes, \( y_{i,c} \) is the ground-truth label, and \( p_{i,c} \) is the predicted probability. For semi-supervised training, a weighted loss incorporates pseudo-labels (weight = 0.8), as detailed in Algorithm 1.

\begin{figure}[t]
    \centering
    \begin{minipage}{0.2\textwidth}
        \centering
        \includegraphics[width=0.8\textwidth, height=0.8\textwidth]{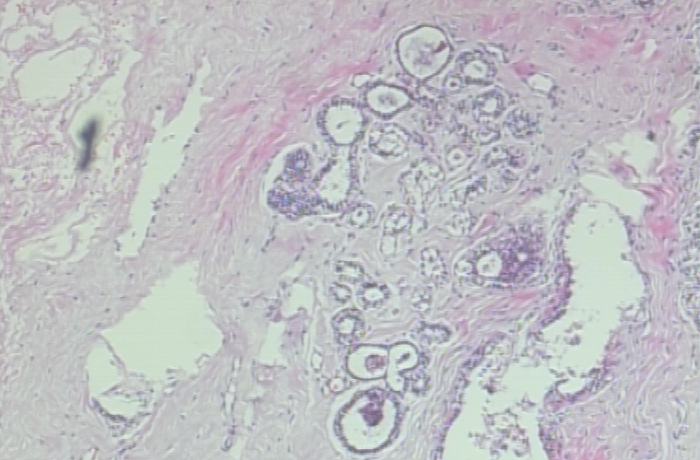}\\
        40x
    \end{minipage}
    \begin{minipage}{0.2\textwidth}
        \centering
        \includegraphics[width=0.8\textwidth, height=0.8\textwidth]{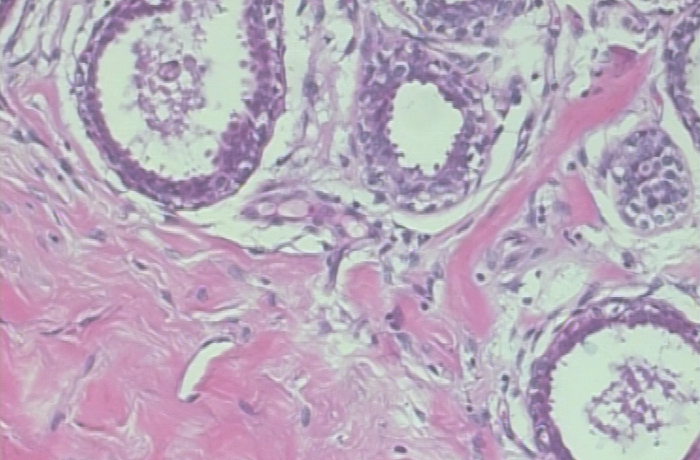}\\
        100x
    \end{minipage}
    
    \vspace{1em}
    
    \begin{minipage}{0.2\textwidth}
        \centering
        \includegraphics[width=0.8\textwidth, height=0.8\textwidth]{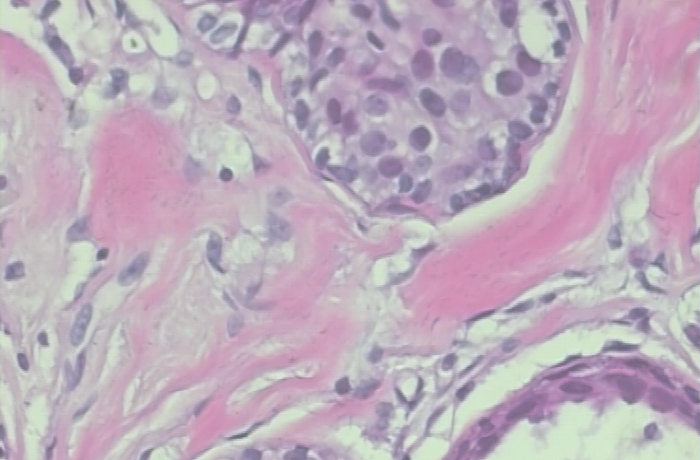}\\
        200x
    \end{minipage}
    \begin{minipage}{0.2\textwidth}
        \centering
        \includegraphics[width=0.8\textwidth, height=0.8\textwidth]{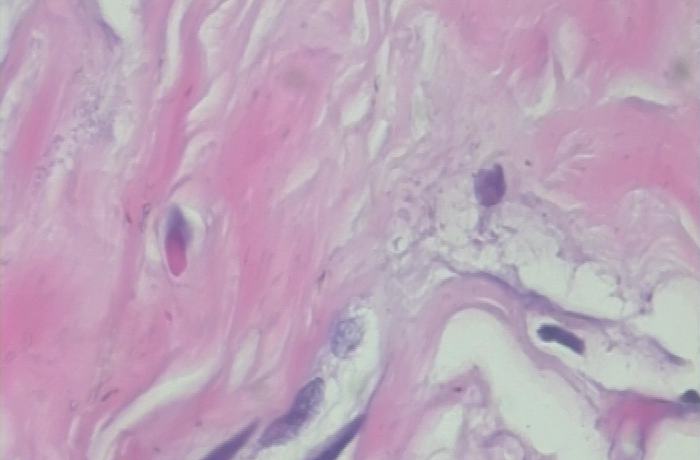}\\
        400x
    \end{minipage}
    
    \caption{Images of Breast Cancer Histopathology from the BreakHis Dataset at Four Magnifications}
    \label{Tal6}
\end{figure}

\begin{figure}[t]
    \centering
    \begin{minipage}{0.2\textwidth}
        \centering
        \includegraphics[width=0.8\textwidth, height=0.8\textwidth]{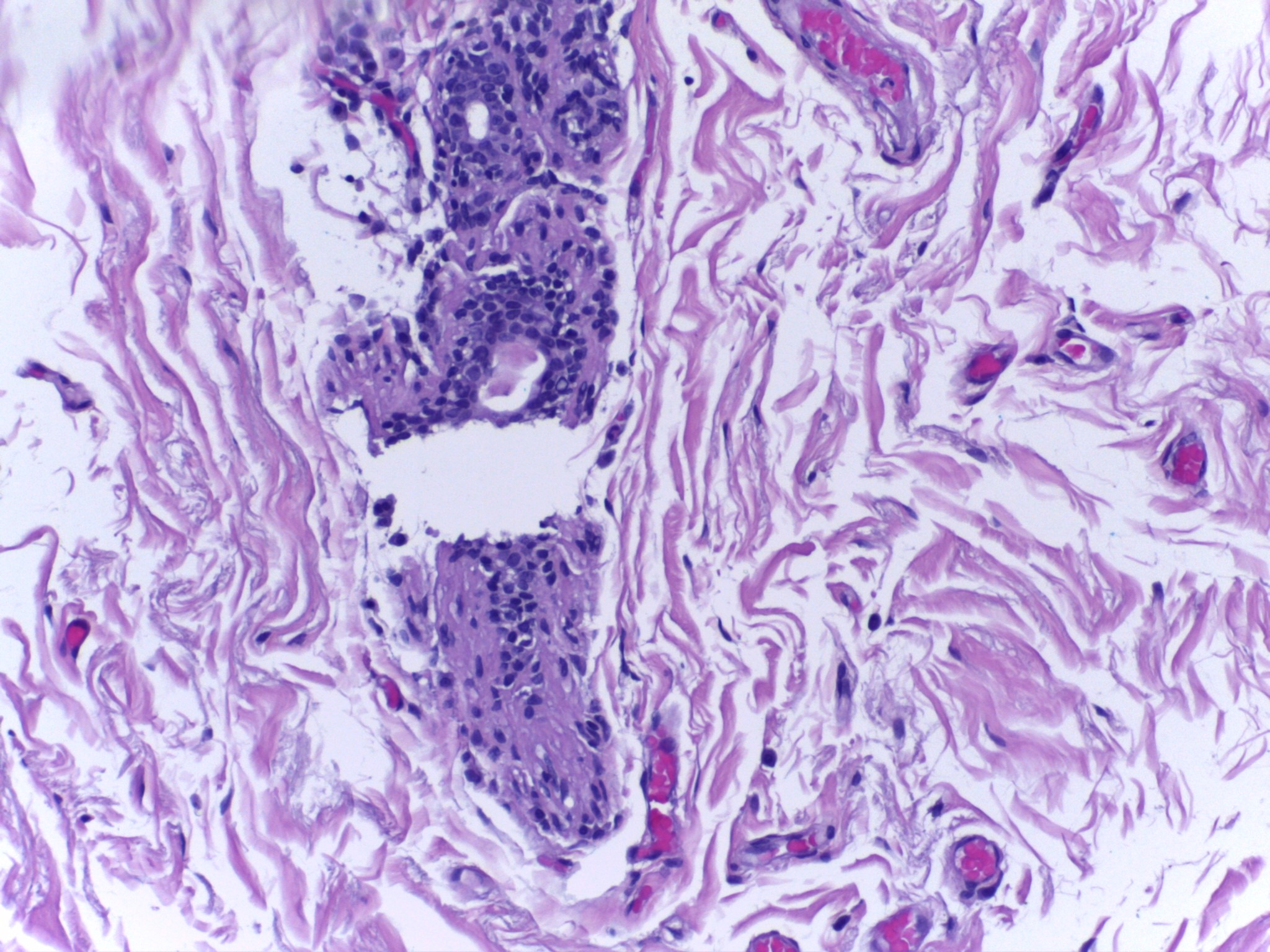}\\
        Normal
    \end{minipage}
    \begin{minipage}{0.2\textwidth}
        \centering
        \includegraphics[width=0.8\textwidth, height=0.8\textwidth]{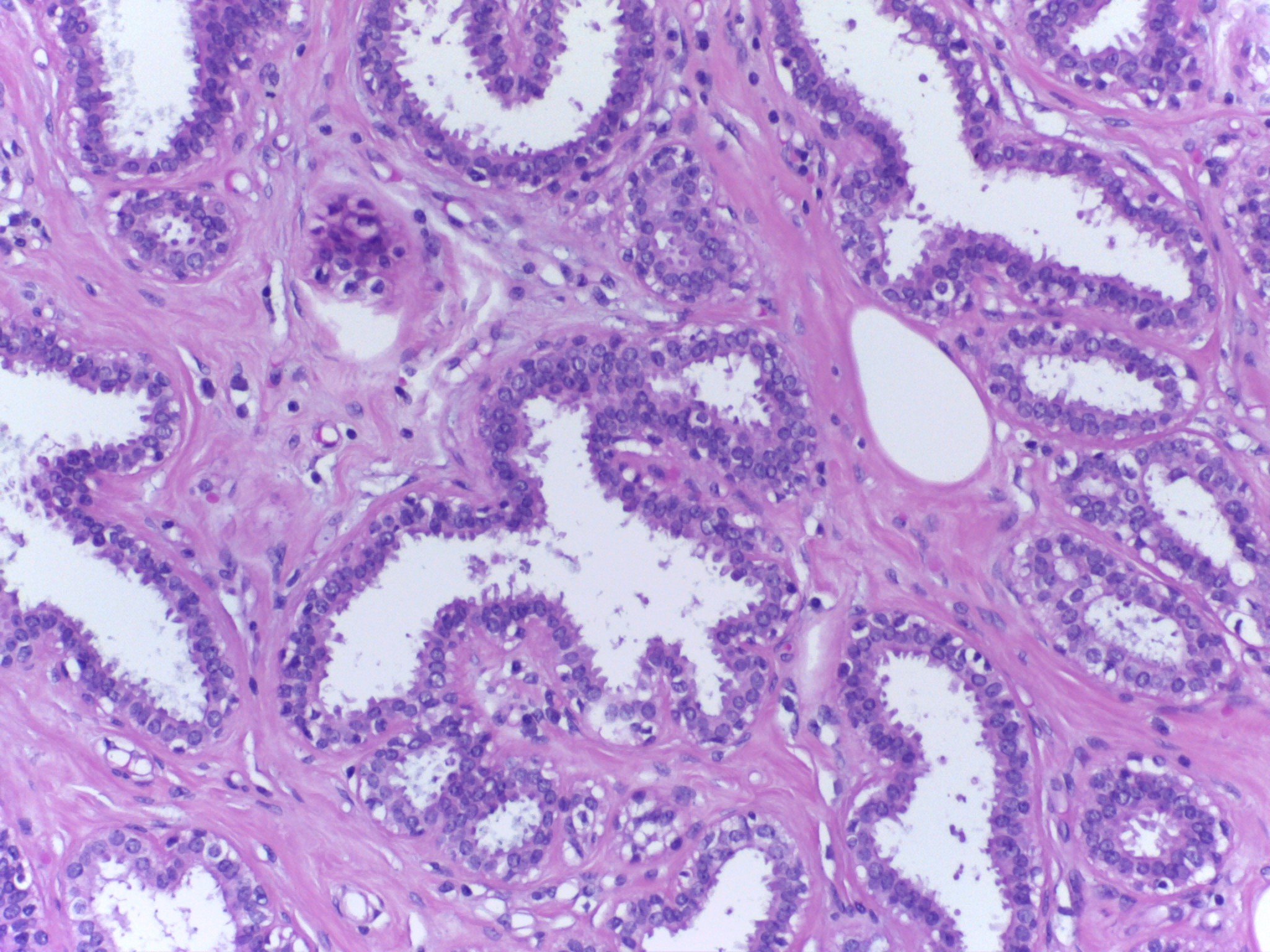}\\
        Benign
    \end{minipage}
    
    \vspace{1em}
    
    \begin{minipage}{0.2\textwidth}
        \centering
        \includegraphics[width=0.8\textwidth, height=0.8\textwidth]{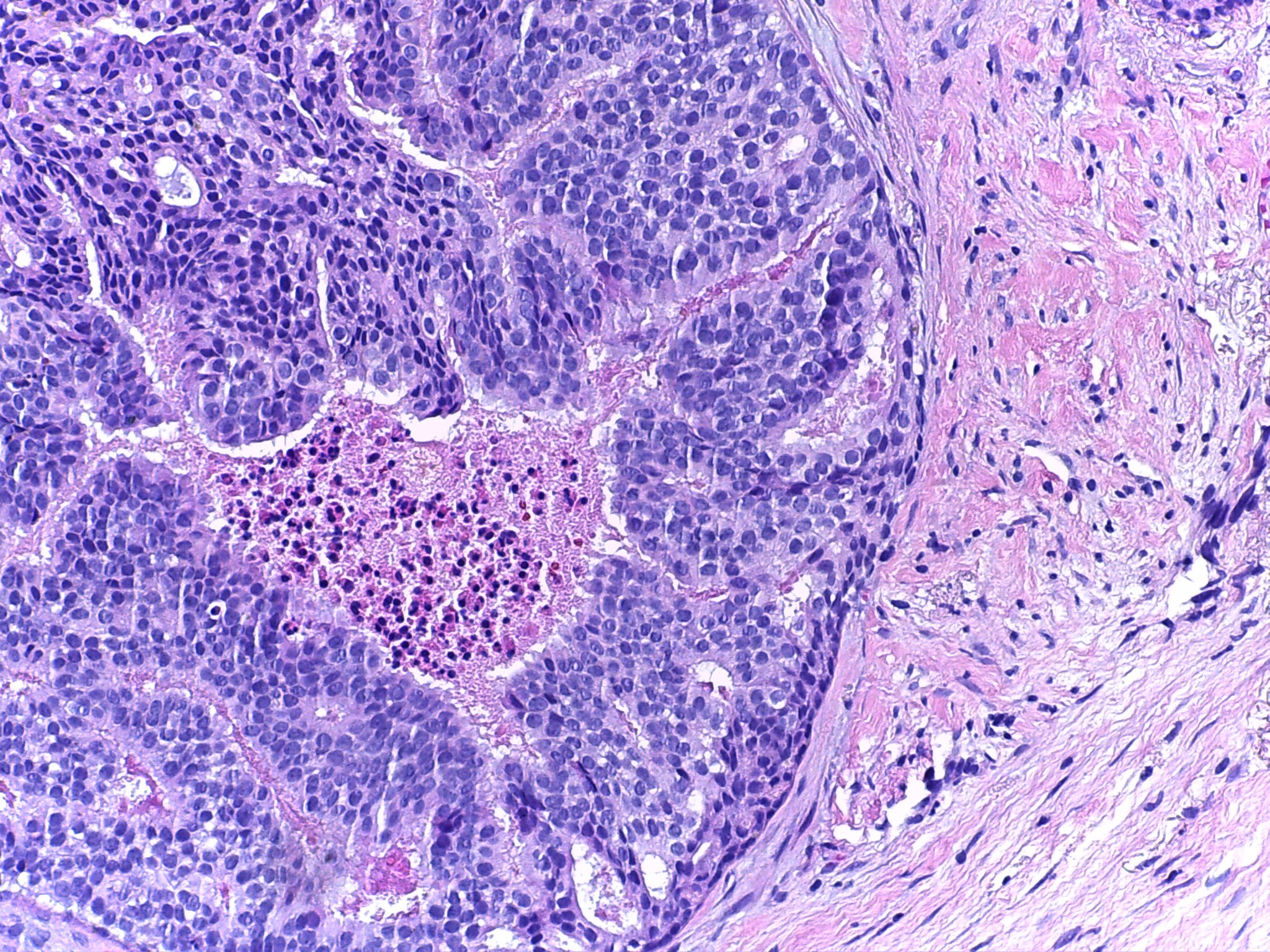}\\
        InSitu
    \end{minipage}
    \begin{minipage}{0.2\textwidth}
        \centering
        \includegraphics[width=0.8\textwidth, height=0.8\textwidth]{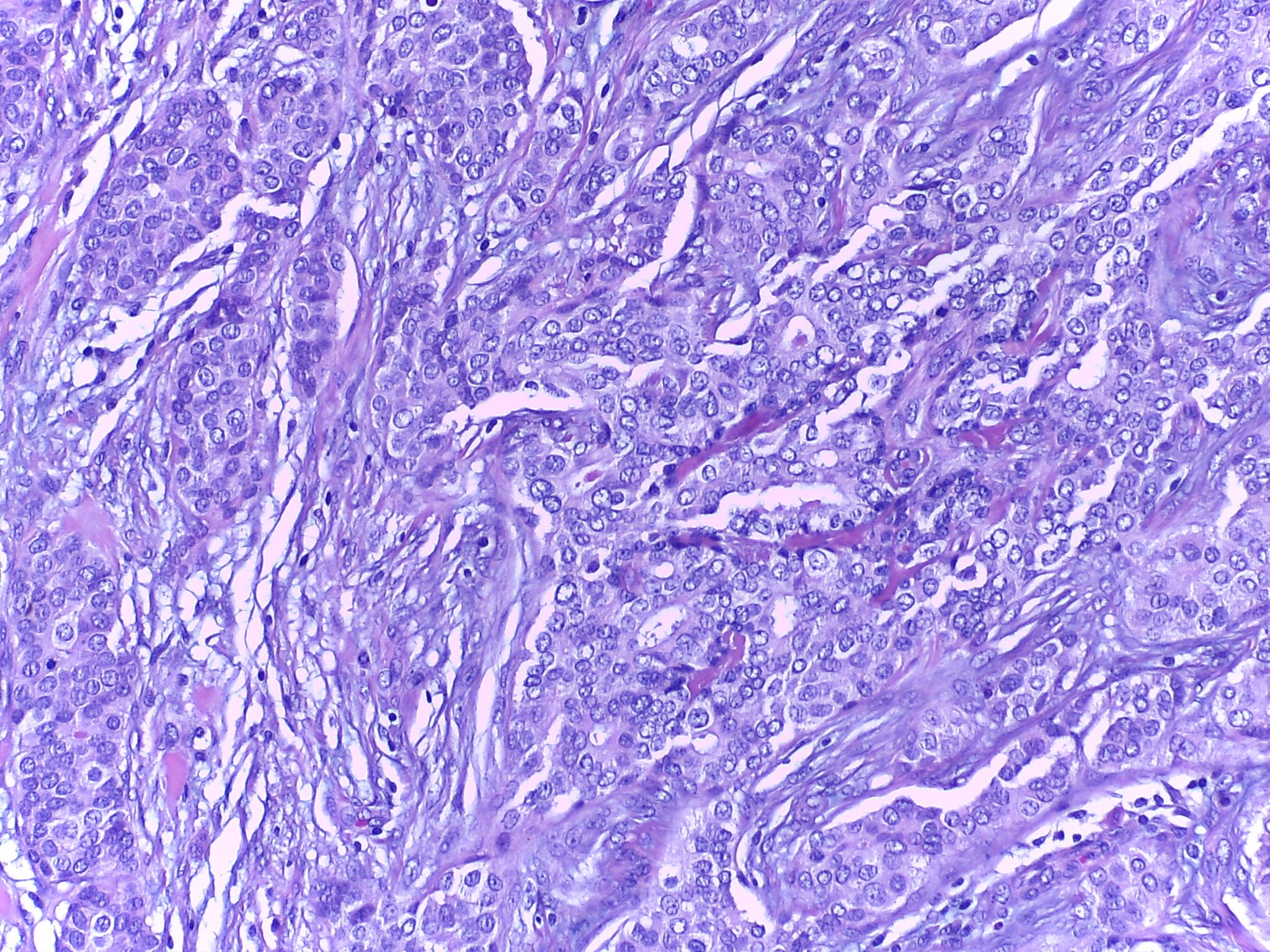}\\
        Invasive
    \end{minipage}

    \caption{Images of Breast Cancer Histopathology from the  ICIAR2018  Dataset }
    \label{Tal6}
\end{figure}

\begin{figure}[h]
    \centering
    \begin{minipage}{0.2\textwidth}
        \centering
        \includegraphics[width=0.8\textwidth, height=0.8\textwidth]{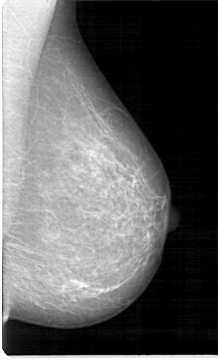}\\
        Normal 
    \end{minipage}
    \begin{minipage}{0.2\textwidth}
        \centering
        \includegraphics[width=0.8\textwidth, height=0.8\textwidth]{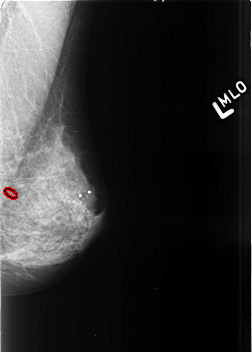}\\
        Benign
    \end{minipage}
    
    \vspace{1em}
    
    \begin{minipage}{0.2\textwidth}
        \centering
        \includegraphics[width=0.8\textwidth, height=0.8\textwidth]{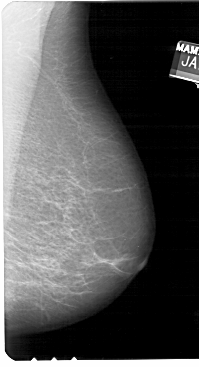}\\
        Malignant
    \end{minipage}

    \caption{Images of Breast Cancer Histopathology from the Mini-DDSM Dataset}
    \label{Tal6}
\end{figure}
\begin{table}[H] 
    \caption{BreakHis dataset: benign and malignant image distribution across magnification factors}
    \label{tab:breakhis_distribution}
    \centering
    \begin{tabular}{m{110pt}<{\centering} | cccc}
         \hline
         \multirow{2}{*}{\textbf{Class}} & \multicolumn{4}{c}{Magnification Factor} \\
         \cline{2-5}
         & \textbf{40X} & \textbf{100X} & \textbf{200X} & \textbf{400X} \\
         \hline
         Benign & 625 & 644 & 623 & 588 \\
         Malignant & 1370 & 1437 & 1390 & 1232 \\ 
         Total No of Images & 1995 & 2081 & 2013 & 1820 \\ 
         \hline
    \end{tabular}
\end{table}

\begin{table}[H]
    \centering
    \caption{BACH Dataset Details}
    \label{tab1}
    \begin{tabular}{c|c|c}
    \hline
    \textbf{Category} & \textbf{Number of Images} & \textbf{Image Size} \\
    \hline
    Normal            & 100                       & 2048 × 1536         \\
     \hline
    Benign            & 100                       & 2048 × 1536         \\
     \hline
    In-situ carcinoma & 100                       & 2048 × 1536         \\
     \hline
    Invasive carcinoma& 100                       & 2048 × 1536         \\
    \hline
    Total  & 400             & 2048 × 1536 \\
    \hline
    \end{tabular}
\end{table}

 \begin{table}[H]
    \centering
    \caption{Detailed Information of the Mini-DDSM Dataset}
    \label{tab:mini_ddsm}
    \begin{tabular}{l|c|l}
    \hline
    \textbf{Category}       & \textbf{Number of Images} & \textbf{Image Size / Format} \\
    \hline
   
    Normal Images           & 500        & Variable                    \\
    \hline
    Benign Images           & 700        & Variable                    \\
    \hline
    Malignant Images        & 800        & Variable                    \\
    \hline
    Total Images            &  2000       & Variable                    \\
    \hline
    \end{tabular}
\end{table}

\begin{table*}[t]
\centering
\caption{Performance of Network Models on BreakHis Dataset (Two-Class Classification) }
\label{tab:performance_40x_100x}
\scriptsize 
\setlength{\tabcolsep}{4pt} 
\resizebox{\textwidth}{!}{ 
\begin{tabular}{cccccccccc}
\hline
\textbf{Network Models} & \multicolumn{4}{c}{\textbf{40X}} & \multicolumn{4}{c}{\textbf{100X}} \\ 
\cline{2-9}
 & \textbf{AUC-ROC} & \textbf{Balanced Accuracy} & \textbf{F1 Score} & \textbf{Cohen's Kappa} & \textbf{AUC-ROC} & \textbf{Balanced Accuracy} & \textbf{F1 Score} & \textbf{Cohen's Kappa} \\ 
\hline
SimCLR \cite{b21} & 0.7956$\pm$0.0155 & 0.7324$\pm$0.0106 & 0.8283$\pm$0.0240 & 0.4631$\pm$0.0306 & 0.8022$\pm$0.0383 & 0.7080$\pm$0.0720 & 0.8144$\pm$0.0090 & 0.4221$\pm$0.1288 \\ 
BYOL \cite{b23} & 0.8645$\pm$0.0159 & 0.7688$\pm$0.0269 & 0.8643$\pm$0.0087 & 0.5483$\pm$0.0437 & 0.8702$\pm$0.0090 & 0.7728$\pm$0.0147 & 0.8681$\pm$0.0073 & 0.5571$\pm$0.0267 \\ 
ViT \cite{b24} & 0.8365$\pm$0.0116 & 0.7423$\pm$0.0390 & 0.8573$\pm$0.0096 & 0.5039$\pm$0.0565 & 0.8676$\pm$0.0212 & 0.7353$\pm$0.0439 & 0.8788$\pm$0.0044 & 0.5215$\pm$0.0598 \\ 
CoAtNet \cite{b26}& 0.9123$\pm$0.0076 & 0.8377$\pm$0.0136 & 0.8449$\pm$0.0112 & 0.6903$\pm$0.0222 & 0.9514$\pm$0.0136 & 0.8710$\pm$0.0113 & 0.8703$\pm$0.0100 & 0.7415$\pm$0.0191 \\ 
DenTnet \cite{b27}& 0.8927$\pm$0.0173 & 0.8146$\pm$0.0067 & 0.8260$\pm$0.0073 & 0.6526$\pm$0.0147 & 0.9325$\pm$0.0047 & 0.8432$\pm$0.0049 & 0.8517$\pm$0.0014 & 0.7037$\pm$0.0031 \\ 
MedViT\cite{b28} & 0.8434$\pm$0.0257 & 0.7909$\pm$0.0206 & 0.7928$\pm$0.0170 & 0.5857$\pm$0.0337 & 0.8368$\pm$0.0562 & 0.6352$\pm$0.1359 & 0.5963$\pm$0.1881 & 0.2851$\pm$0.2860 \\ 
EfficientNetV2 \cite{b29}& 0.8138$\pm$0.0148 & 0.7543$\pm$0.0194 & 0.7701$\pm$0.0195 & 0.5437$\pm$0.0383 & 0.8666$\pm$0.0167 & 0.7537$\pm$0.0164 & 0.7711$\pm$0.0145 & 0.5462$\pm$0.0276 \\ 
SWIN-UNETR \cite{b30}& 0.9160$\pm$0.0133 & 0.8229$\pm$0.0183 & 0.8309$\pm$0.0169 & 0.6623$\pm$0.0335 & 0.9473$\pm$0.0028 & 0.8792$\pm$0.0076 & 0.8798$\pm$0.0060 & 0.7597$\pm$0.0121 \\ 
MLFF \cite{b31} & 0.9008$\pm$0.0076 & 0.8133$\pm$0.0055 & 0.8291$\pm$0.0042 & 0.6597$\pm$0.0080 & 0.9586$\pm$0.0021 & 0.8707$\pm$0.0185 & 0.8862$\pm$0.0178 & 0.7731$\pm$0.0354 \\ 
\textbf{DCS-ST} & \textbf{0.9187$\pm$0.0051} & \textbf{0.8664$\pm$0.0053} & \textbf{0.9004$\pm$0.0126} & \textbf{0.7695$\pm$0.0244} & \textbf{0.9614$\pm$0.0182} & \textbf{0.8833$\pm$0.0303} & \textbf{0.9217$\pm$0.0144} & \textbf{0.7884$\pm$0.0531} \\ 
\hline
\end{tabular}
}
\end{table*}

\vspace{0.3cm} 
\begin{table*}[t]
\centering
\label{tab:performance_200x_400x}
\scriptsize 
\setlength{\tabcolsep}{4pt} 
\resizebox{\textwidth}{!}{ 
\begin{tabular}{cccccccccc}
\hline
\textbf{Network Models} & \multicolumn{4}{c}{\textbf{200X}} & \multicolumn{4}{c}{\textbf{400X}} \\ 
\cline{2-9}
 & \textbf{AUC-ROC} & \textbf{Balanced Accuracy} & \textbf{F1 Score} & \textbf{Cohen's Kappa} & \textbf{AUC-ROC} & \textbf{Balanced Accuracy} & \textbf{F1 Score} & \textbf{Cohen's Kappa} \\ 
\hline
SimCLR \cite{b21} & 0.8669$\pm$0.0153 & 0.7529$\pm$0.0338 & 0.8708$\pm$0.0074 & 0.5348$\pm$0.0496 & 0.8355$\pm$0.0277 & 0.7607$\pm$0.0250 & 0.8441$\pm$0.0311 & 0.5237$\pm$0.0646 \\ 
BYOL \cite{b23} & 0.9037$\pm$0.0243 & 0.8088$\pm$0.0325 & 0.8929$\pm$0.0172 & 0.6355$\pm$0.0629 & 0.8175$\pm$0.0131 & 0.7006$\pm$0.0209 & 0.8441$\pm$0.0093 & 0.4364$\pm$0.0410 \\ 
ViT \cite{b24} & 0.9211$\pm$0.0081 & 0.8015$\pm$0.0530 & 0.9075$\pm$0.0161 & 0.6502$\pm$0.0861 & 0.8921$\pm$0.0046 & 0.7903$\pm$0.0528 & 0.8403$\pm$0.0825 & 0.5708$\pm$0.1188 \\ 
CoAtNet\cite{b26} & 0.9471$\pm$0.0078 & 0.8623$\pm$0.0060 & 0.8768$\pm$0.0068 & 0.7542$\pm$0.0138 & 0.8866$\pm$0.0106 & 0.8148$\pm$0.0193 & 0.8305$\pm$0.0177 & 0.6629$\pm$0.0347 \\ 
DenTnet\cite{b27} & 0.9110$\pm$0.0006 & 0.7983$\pm$0.0005 & 0.8158$\pm$0.0029 & 0.6337$\pm$0.0062 & 0.8834$\pm$0.0072 & 0.7742$\pm$0.0004 & 0.7967$\pm$0.0038 & 0.5994$\pm$0.0099 \\ 
MedViT\cite{b28} & 0.8874$\pm$0.0227 & 0.8326$\pm$0.0178 & 0.8470$\pm$0.0137 & 0.6950$\pm$0.0267 & 0.8742$\pm$0.0128 & 0.7653$\pm$0.0555 & 0.7881$\pm$0.0596 & 0.5885$\pm$0.1069 \\ 
EfficientNetV2\cite{b29} & 0.8967$\pm$0.0158 & 0.7995$\pm$0.0126 & 0.8231$\pm$0.0086 & 0.6503$\pm$0.0158 & 0.8619$\pm$0.0149 & 0.7732$\pm$0.0210 & 0.7889$\pm$0.0200 & 0.5811$\pm$0.0391 \\ 
SWIN-UNETR\cite{b30} & 0.9411$\pm$0.0078 & 0.8830$\pm$0.0170 & 0.8893$\pm$0.0146 & 0.7787$\pm$0.0291 & 0.8611$\pm$0.0177 & 0.7818$\pm$0.0160 & 0.8020$\pm$0.0142 & 0.6085$\pm$0.0265 \\ 
MLFF\cite{b31} & 0.9478$\pm$0.0042 & 0.8772$\pm$0.0116 & 0.8747$\pm$0.0118 & 0.7494$\pm$0.0236 & 0.8727$\pm$0.0140 & 0.7822$\pm$0.0179 & 0.7926$\pm$0.0189 & 0.5865$\pm$0.0378 \\ 
\textbf{DCS-ST} & \textbf{0.9528$\pm$0.0026} & \textbf{0.8935$\pm$0.0084} & \textbf{0.9339$\pm$0.0039} & \textbf{0.7802$\pm$0.0141} & \textbf{0.8945$\pm$0.0243} & \textbf{0.8263$\pm$0.0461} & \textbf{0.8805$\pm$0.0120} & \textbf{0.6985$\pm$0.0796} \\ 
\hline
\end{tabular}
}
\end{table*}

\begin{table*}[t]
\centering
\caption{Performance of Network Models on Mini-DDSM Dataset (Three-Class Classification) and ICIAR2018 Dataset (Four-Class Classification)}
\label{your_table_label}
\scriptsize 
\setlength{\tabcolsep}{4pt} 
\resizebox{\textwidth}{!}{ 
\begin{tabular}{cccccccccc}
\hline
\textbf{Network Models} & \multicolumn{4}{c}{\textbf{Mini-DDSM Dataset (Three-Class Classification)}} & \multicolumn{4}{c}{\textbf{ICIAR2018 Dataset (Four-Class Classification)}} \\ 
\cline{2-9}
 & \textbf{AUC-ROC} & \textbf{Balanced Accuracy} & \textbf{F1 Score} & \textbf{Cohen's Kappa} & \textbf{AUC-ROC} & \textbf{Balanced Accuracy} & \textbf{F1 Score} & \textbf{Cohen's Kappa} \\ 
\hline
SimCLR \cite{b21} & 0.6687$\pm$0.0420 & 0.6089$\pm$0.0404 & 0.4740$\pm$0.0487 & 0.2096$\pm$0.0775 & 0.6356$\pm$0.0068 & 0.5875$\pm$0.0191 & 0.3773$\pm$0.0209 & 0.1750$\pm$0.0382 \\ 
BYOL \cite{b23} & 0.7686$\pm$0.0029 & 0.5882$\pm$0.0039 & 0.5779$\pm$0.0043 & 0.3632$\pm$0.0062 & 0.7584$\pm$0.0346 & 0.5094$\pm$0.0503 & 0.5073$\pm$0.0471 & 0.3458$\pm$0.0671 \\ 
ViT \cite{b24} & 0.7447$\pm$0.0115 & 0.6702$\pm$0.0228 & 0.5251$\pm$0.0412 & 0.3217$\pm$0.0466 & 0.7721$\pm$0.0060 & 0.6896$\pm$0.0216 & 0.5044$\pm$0.0406 & 0.3792$\pm$0.0431 \\ 
CoAtNet \cite{b26} & 0.8168$\pm$0.0089 & 0.7323$\pm$0.0076 & 0.6432$\pm$0.0121 & 0.4542$\pm$0.0160 & 0.6978$\pm$0.0303 & 0.6125$\pm$0.0320 & 0.4116$\pm$0.0508 & 0.2250$\pm$0.0640 \\
DenTnet \cite{b27}& 0.8049$\pm$0.0093 & 0.7224$\pm$0.0102 & 0.6284$\pm$0.0113 & 0.4345$\pm$0.0189 & 0.6978$\pm$0.0303 & 0.6125$\pm$0.0320 & 0.4116$\pm$0.0508 & 0.2250$\pm$0.0640 \\
MedViT \cite{b28}& 0.7098$\pm$0.0559 & 0.6377$\pm$0.0460 & 0.4781$\pm$0.0676 & 0.2603$\pm$0.0881 & 0.8161$\pm$0.0310 & 0.7354$\pm$0.0345 & 0.5996$\pm$0.0571 & 0.4708$\pm$0.0691 \\
EfficientNetV2 \cite{b29} & 0.7251$\pm$0.0033 & 0.6515$\pm$0.0062 & 0.5033$\pm$0.0065 & 0.2859$\pm$0.0102 & 0.5509$\pm$0.0494 & 0.5104$\pm$0.0180 & 0.1263$\pm$0.0456 & 0.0208$\pm$0.0361 \\
SWIN-UNETR \cite{b30} & 0.8325$\pm$0.0293 & 0.7376$\pm$0.0282 & 0.6476$\pm$0.0385 & 0.4645$\pm$0.0570 & 0.8726$\pm$0.0341 & 0.7708$\pm$0.0273 & 0.6601$\pm$0.0378 & 0.5417$\pm$0.0546 \\
MLFF\cite{b31} & 0.8262$\pm$0.0045 & 0.7397$\pm$0.0014 & 0.6556$\pm$0.0020 & 0.4711$\pm$0.0025 & 0.8190$\pm$0.0256 & 0.7500$\pm$0.0243 & 0.6237$\pm$0.0356 & 0.5000$\pm$0.0486 \\
\textbf{DCS-ST} & \textbf{0.8334$\pm$0.0523} & \textbf{0.7541$\pm$0.0338} & \textbf{0.6687$\pm$0.0657} & \textbf{0.6355$\pm$0.0722} & \textbf{0.8781$\pm$0.0201} & \textbf{0.7771$\pm$0.0207} & \textbf{0.6712$\pm$0.0368} & \textbf{0.6542$\pm$0.0415} \\
\hline
\end{tabular}
}
\end{table*}

\section{Experiment}
\subsection{ Experimental Settings}
\subsubsection{Implementation Details}

Our implementation is built on PyTorch 2.0.1 and Python 3.10, with training conducted on two NVIDIA T4 GPUs (Kaggle T4*2 setup), each with 16GB of video memory. To handle the small and imbalanced medical imaging dataset, we set a conservative initial learning rate of 0.0001, mitigating the risk of loss explosion and ensuring stable convergence. Due to memory constraints, the batch size is set to 16 (as defined in the DataLoader), enabling frequent parameter updates. The model is trained for 50 epochs, balancing training effectiveness and computational efficiency. To ensure robust evaluation, we conduct four independent experiments, averaging their results to account for variability and enhance the reliability of our findings.

\begin{table*}[t]
\centering
\caption{Ablation Experiment: Performance Comparison of Swin-Transformer and DCS-ST on BreakHis Dataset Based on Different Magnifications (Two-Class Classification)}
\label{table_magnification_40x_100x}
\scriptsize
\setlength{\tabcolsep}{4pt}
\resizebox{\textwidth}{!}{
\begin{tabular}{ccccccccc}
\hline
\textbf{Network Models} & \multicolumn{4}{c}{\textbf{40X}} & \multicolumn{4}{c}{\textbf{100X}} \\
\cline{2-9}
 & \textbf{AUC-ROC} & \textbf{Balanced Accuracy} & \textbf{F1 Score} & \textbf{Cohen's Kappa} & \textbf{AUC-ROC} & \textbf{Balanced Accuracy} & \textbf{F1 Score} & \textbf{Cohen's Kappa} \\
\hline
Swin-Transformer \cite{b25} & $0.9143 \pm 0.0078$ & $0.8564 \pm 0.0073$ & $0.8990 \pm 0.0079$ & $0.6942 \pm 0.0171$ & $0.9315 \pm 0.0033$ & $0.8686 \pm 0.0171$ & $0.9189 \pm 0.0049$ & $0.7372 \pm 0.0229$ \\
\textbf{DCS-ST} & $\mathbf{0.9187} \pm \mathbf{0.0051}$ & $\mathbf{0.8664} \pm \mathbf{0.0053}$ & $\mathbf{0.9004} \pm \mathbf{0.0126}$ & $\mathbf{0.7695} \pm \mathbf{0.0244}$ & $\mathbf{0.9614} \pm \mathbf{0.0182}$ & $\mathbf{0.8833} \pm \mathbf{0.0303}$ & $\mathbf{0.9217} \pm \mathbf{0.0144}$ & $\mathbf{0.7884} \pm \mathbf{0.0531}$ \\
\hline
\end{tabular}
}
\end{table*}

\vspace{0.3cm}

\begin{table*}[t]
\centering
\caption{Continuation: Performance Comparison of Swin-Transformer and DCS-ST on BreakHis Dataset (200X and 400X)}
\label{table_magnification_200x_400x}
\scriptsize
\setlength{\tabcolsep}{4pt}
\resizebox{\textwidth}{!}{
\begin{tabular}{ccccccccc}
\hline
\textbf{Network Models} & \multicolumn{4}{c}{\textbf{200X}} & \multicolumn{4}{c}{\textbf{400X}} \\
\cline{2-9}
 & \textbf{AUC-ROC} & \textbf{Balanced Accuracy} & \textbf{F1 Score} & \textbf{Cohen's Kappa} & \textbf{AUC-ROC} & \textbf{Balanced Accuracy} & \textbf{F1 Score} & \textbf{Cohen's Kappa} \\
\hline
Swin-Transformer \cite{b25} & $0.9430 \pm 0.0027$ & $0.8786 \pm 0.0049$ & $0.9313 \pm 0.0026$ & $0.7711 \pm 0.0090$ & $0.8702 \pm 0.0482$ & $0.7678 \pm 0.0581$ & $0.8519 \pm 0.0278$ & $0.5355 \pm 0.0982$ \\
\textbf{DCS-ST} & $\mathbf{0.9528} \pm \mathbf{0.0026}$ & $\mathbf{0.8935} \pm \mathbf{0.0084}$ & $\mathbf{0.9339} \pm \mathbf{0.0039}$ & $\mathbf{0.7802} \pm \mathbf{0.0141}$ & $\mathbf{0.8945} \pm \mathbf{0.0243}$ & $\mathbf{0.8263} \pm \mathbf{0.0461}$ & $\mathbf{0.8805} \pm \mathbf{0.0120}$ & $\mathbf{0.6985} \pm \mathbf{0.0796}$ \\
\hline
\end{tabular}
}
\end{table*}

\begin{table*}[t]
\centering
\caption{Ablation Experiment: Performance Comparison of Swin-Transformer and DCS-ST on BreakHis Dataset Based on  BACH Dataset (Four-Class Classification) and Mini-DDSM Dataset (Three-Class Classification)}
\label{your_table_label}
\scriptsize 
\setlength{\tabcolsep}{4pt} 
\resizebox{\textwidth}{!}{ 
\begin{tabular}{cccccccccc}
\hline
\textbf{Network Models} & \multicolumn{4}{c}{\textbf{Mini-DDSM Dataset (Three-Class Classification)}} & \multicolumn{4}{c}{\textbf{ICIAR2018 Dataset (Four-Class Classification)}} \\ 
\cline{2-9}
 & \textbf{AUC-ROC} & \textbf{Balanced Accuracy} & \textbf{F1 Score} & \textbf{Cohen's Kappa} & \textbf{AUC-ROC} & \textbf{Balanced Accuracy} & \textbf{F1 Score} & \textbf{Cohen's Kappa} \\ 
\hline
Swin-Transformer \cite{b25} & 0.7820$\pm$0.0173 & 0.6999$\pm$0.0140 & 0.5972$\pm$0.0137 & 0.3886$\pm$0.0266 & 0.8279$\pm$0.0237 & 0.7375$\pm$0.0325 & 0.6080$\pm$0.0476 & 0.4750$\pm$0.0651 \\ 
\textbf{DCS-ST} & \textbf{0.8334$\pm$0.0523} & \textbf{0.7541$\pm$0.0338} & \textbf{0.6687$\pm$0.0657} & \textbf{0.6355$\pm$0.0722} & \textbf{0.8781$\pm$0.0201} & \textbf{0.7771$\pm$0.0207} & \textbf{0.6712$\pm$0.0368} & \textbf{0.5542$\pm$0.0415} \\
\hline
\end{tabular}
}
\end{table*}

\subsubsection{Dataset}

Our network is trained and evaluated on three main datasets: BreakHis \cite{b6}, containing 7,909 breast cancer histopathology images (2,429 benign and 5,429 malignant) at 40×, 100×, 200×, and 400× magnifications, each 700×460 pixels; Mini-DDSM \cite{b7}, a reduced version of DDSM with approximately 2,000 annotated mammography images detailing lesion locations and diagnoses; and BACH from ICIAR2018 \cite{b8}, with 400 HE-stained images in four categories (Normal, Benign, In-situ carcinoma, Invasive carcinoma), each 2048×1536 pixels.
Fig. 2, 3, and 4 depict sample images from the BreakHis, ICIAR2018, and Mini-DDSM datasets. Tables I, II, and III also present statistical data for these datasets, including sample sizes and class distributions.

\subsubsection{ Metrics}

To comprehensively evaluate our model's performance, we adopt a diverse set of complementary metrics as shown in Table IV. Our evaluation framework includes AUC-ROC [9], which quantifies the model's discriminative ability across classification thresholds; balanced accuracy [10], which equalizes the importance of each class regardless of prevalence; F1 score [11], the harmonic mean of precision and recall that proves particularly valuable for imbalanced datasets like BreakHis; and Cohen's Kappa [12], which assesses agreement between predicted and actual classifications while accounting for chance agreement. This multifaceted approach ensures robust performance assessment, especially critical in medical image analysis, where classification errors carry significant consequences. The combined use of these metrics addresses various aspects of model performance—from general classification accuracy to class-imbalance robustness—providing a holistic view that a single metric alone cannot capture. Moreover, this comprehensive evaluation strategy facilitates fair comparison with existing methodologies in the literature while offering deeper insights into our model's strengths and limitations.

 \subsection{Comparison with Existing Methods}

\subsubsection{Comparison Methods}
We evaluate our proposed DCS-ST model against state-of-the-art methods, including SimCLR \cite{b21}, BYOL\cite{b23}, ViT\cite{b24}, CoAtNet\cite{b26}, DenTnet\cite{b27}, MedViT \cite{b28}, EfficientNetV2\cite{b29}, SWIN-UNETR\cite{b30}, and MLFF\cite{b31}, on the BreakHis(two-class), Mini-DDSM(three-class), and ICIAR2018(four-class) datasets. All models are assessed under identical conditions using metrics such as AUC-ROC, Balanced Accuracy, F1 Score, and Cohen’s Kappa, with results reported as mean $\pm$ standard deviation in Tables  IV and V.

\begin{figure*}[t]
    \centering
    
    \begin{minipage}{0.22\textwidth}
        \centering
\includegraphics[width=\linewidth]{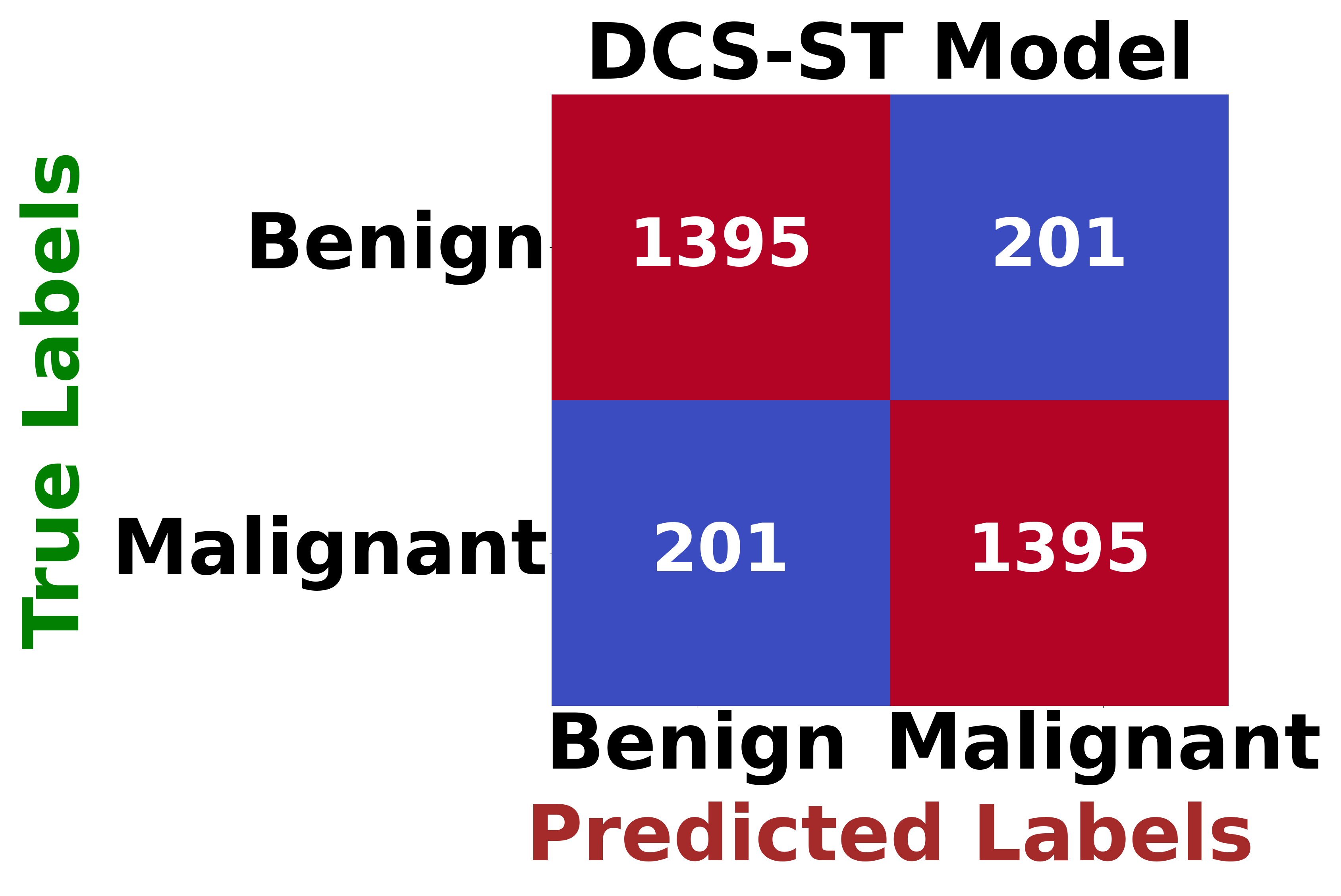}
    \end{minipage}%
    \hspace{0.02\textwidth}
    \begin{minipage}{0.22\textwidth}
        \centering
        \includegraphics[width=\linewidth]{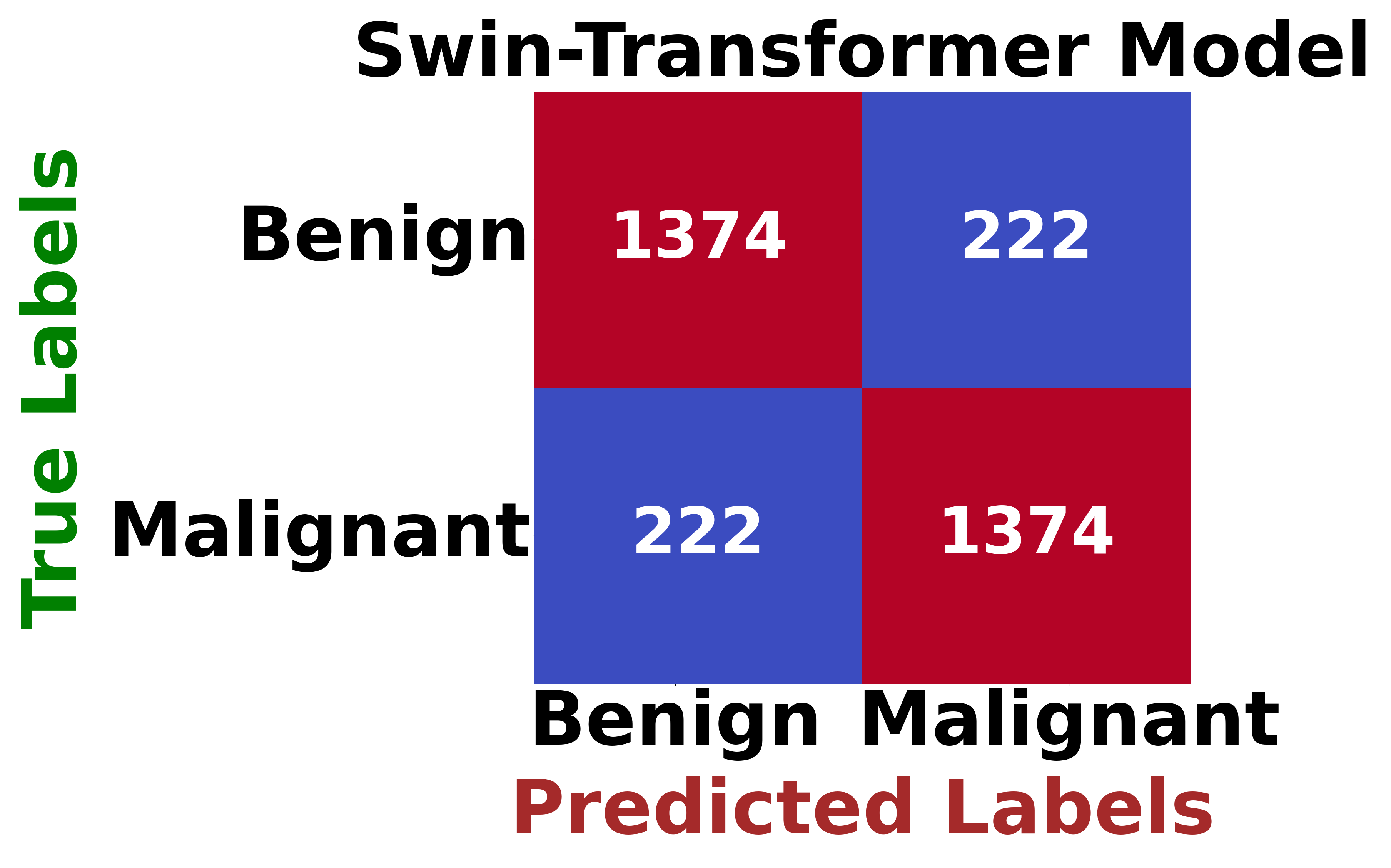}
    \end{minipage}%
    \hspace{0.05\textwidth}
    \begin{minipage}{0.22\textwidth}
        \centering
        \includegraphics[width=\linewidth]{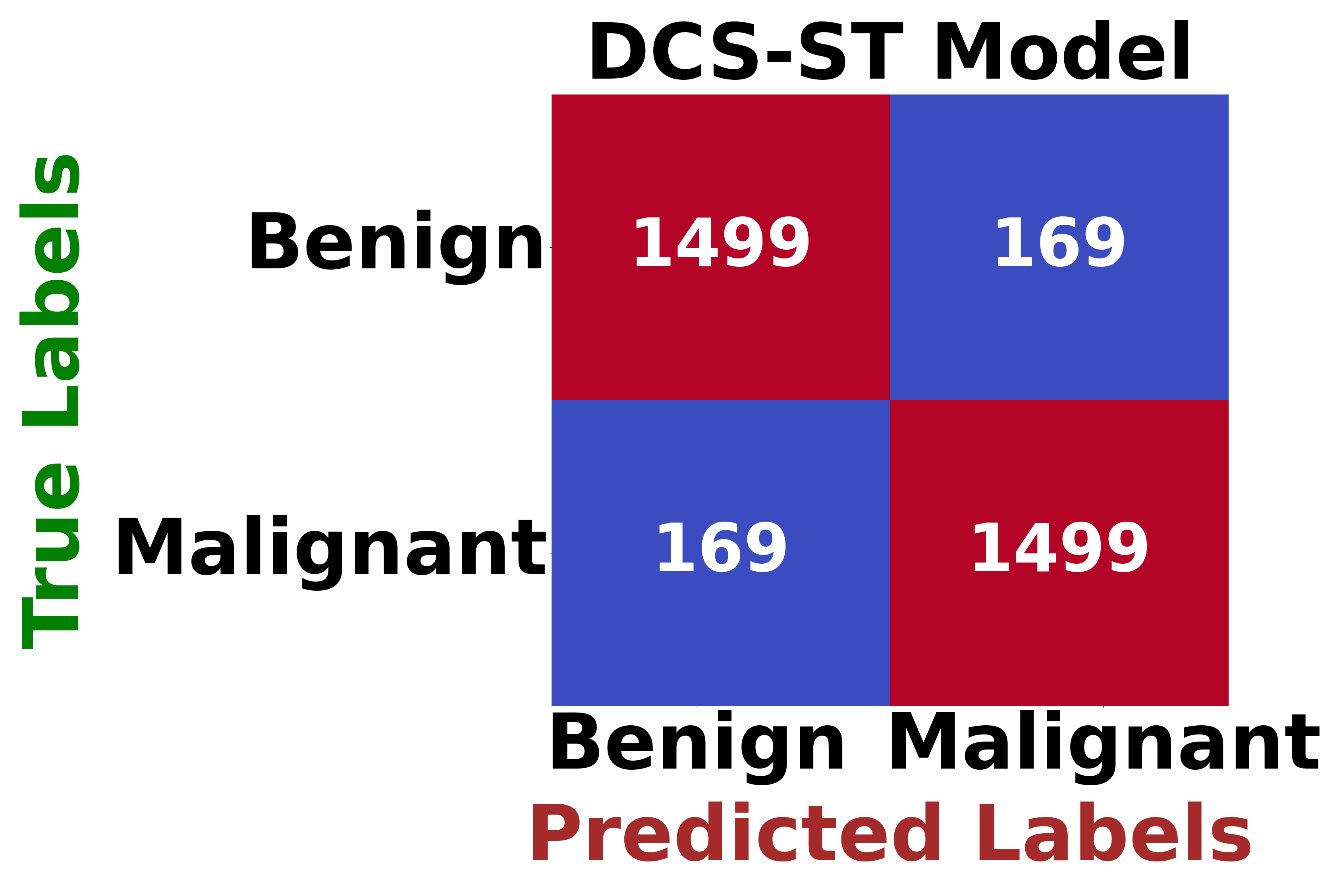}
    \end{minipage}%
    \hspace{0.02\textwidth}
    \begin{minipage}{0.22\textwidth}
        \centering
        \includegraphics[width=\linewidth]{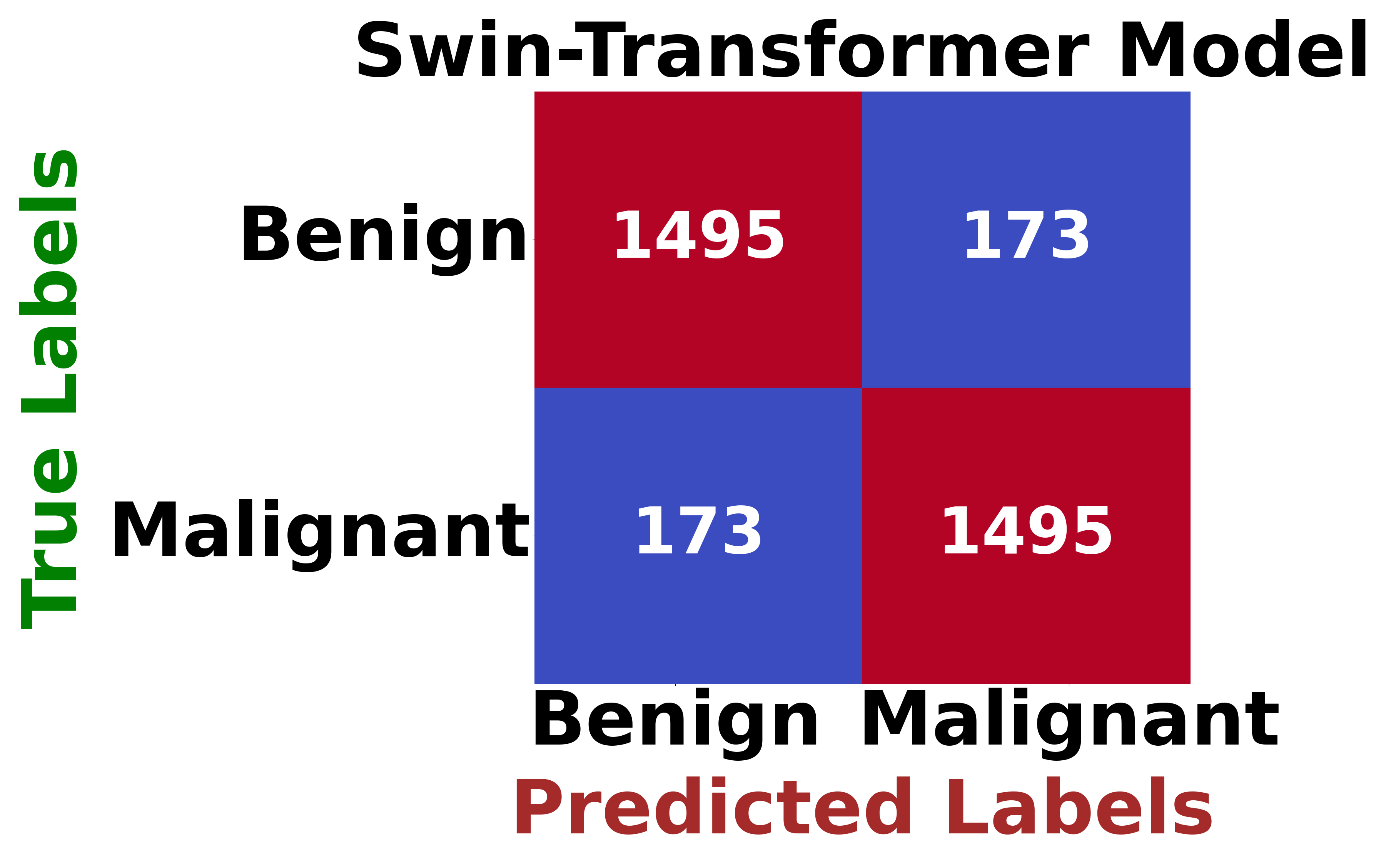}
    \end{minipage}

    \vspace{0.3cm} 
    \begin{minipage}{0.22\textwidth}
        \centering
         40x
    \end{minipage}%
    \hspace{0.29\textwidth} 
    \begin{minipage}{0.22\textwidth}
        \centering
         100x
    \end{minipage}

\end{figure*}
\vspace{0.3cm} 

\begin{figure*}[t]
    \centering
    
    \begin{minipage}{0.22\textwidth}
        \centering
        \includegraphics[width=\linewidth]{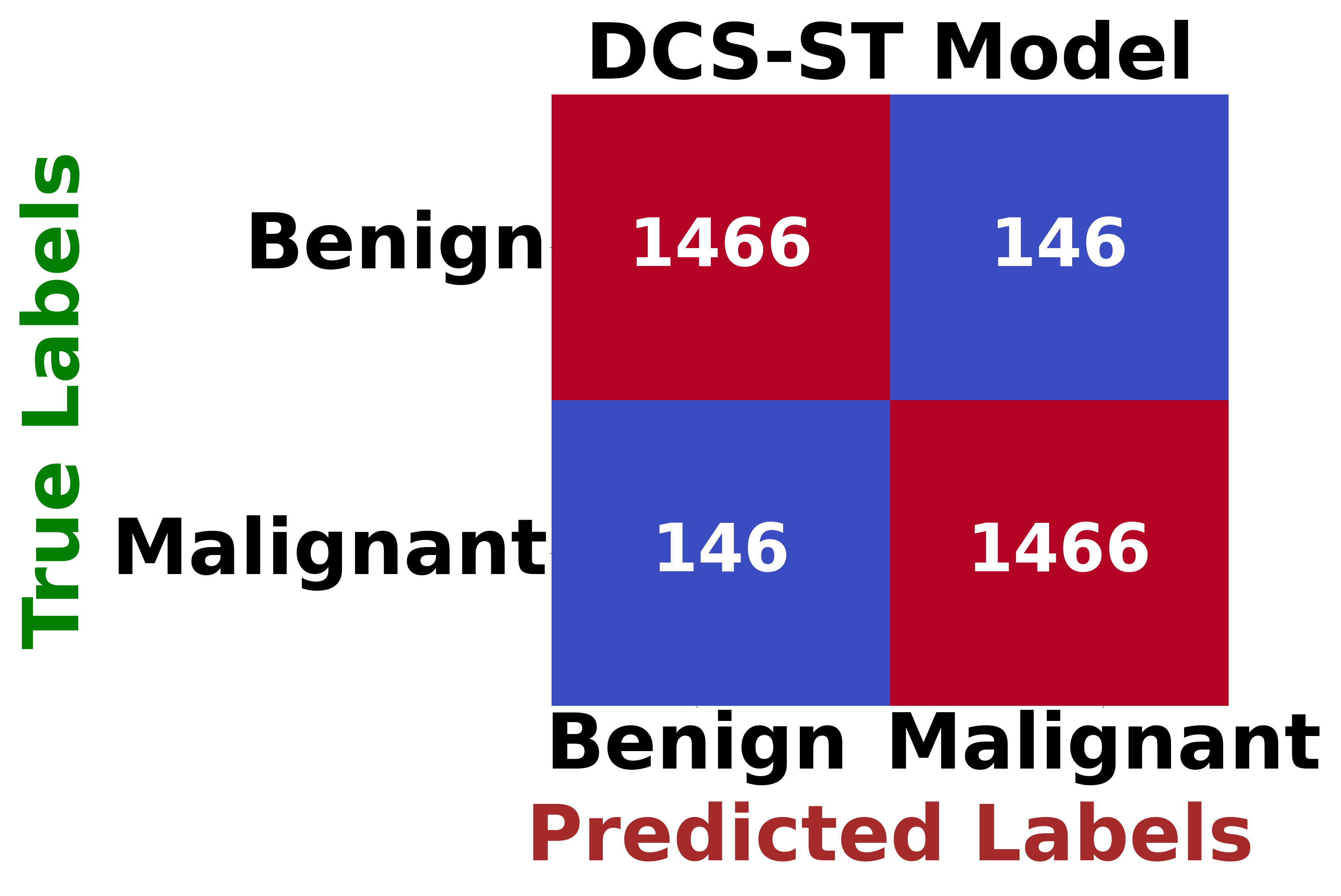}
    \end{minipage}%
    \hspace{0.02\textwidth}
    \begin{minipage}{0.22\textwidth}
        \centering
        \includegraphics[width=\linewidth]{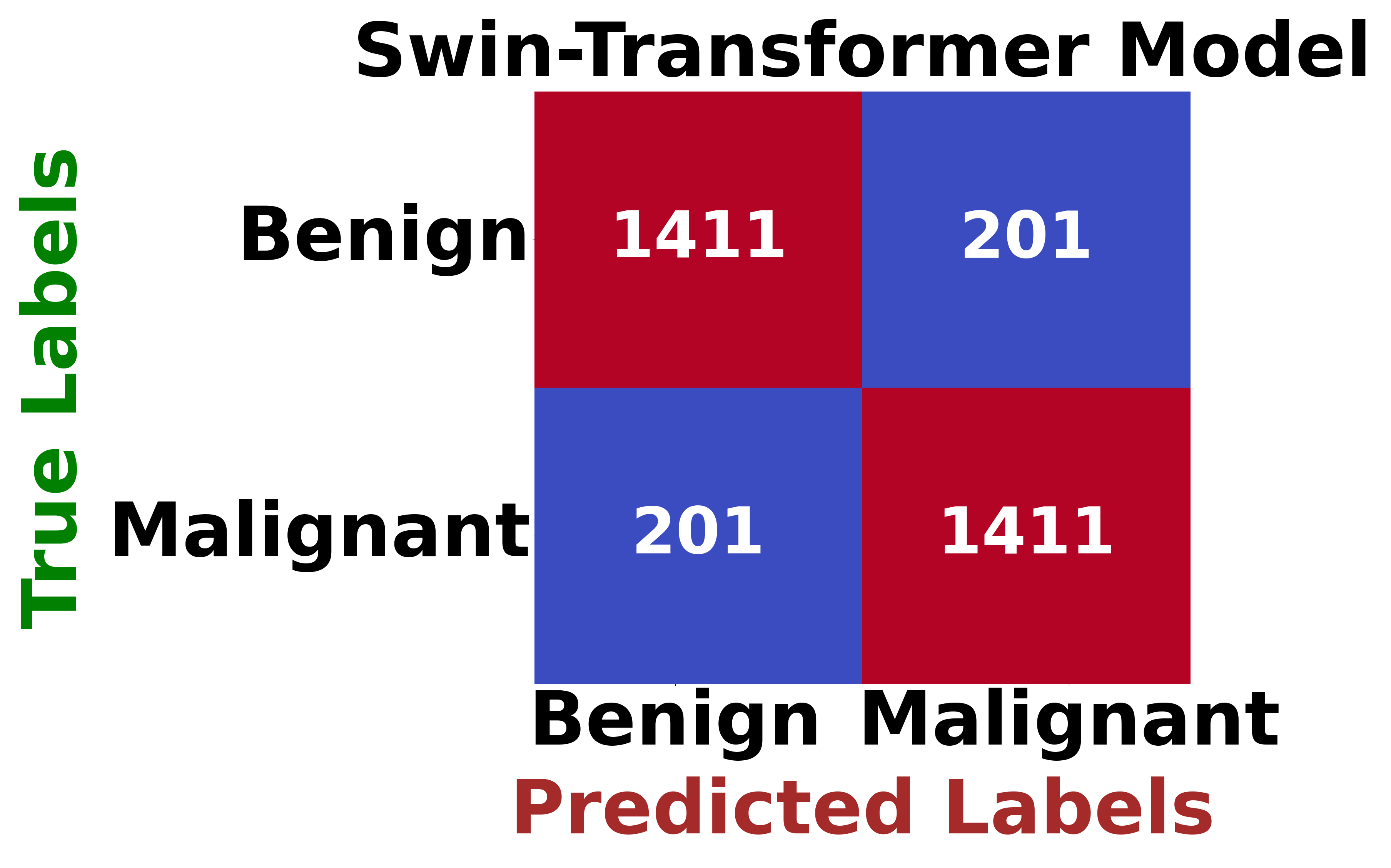}
    \end{minipage}%
    \hspace{0.05\textwidth}
    \begin{minipage}{0.22\textwidth}
        \centering
        \includegraphics[width=\linewidth]{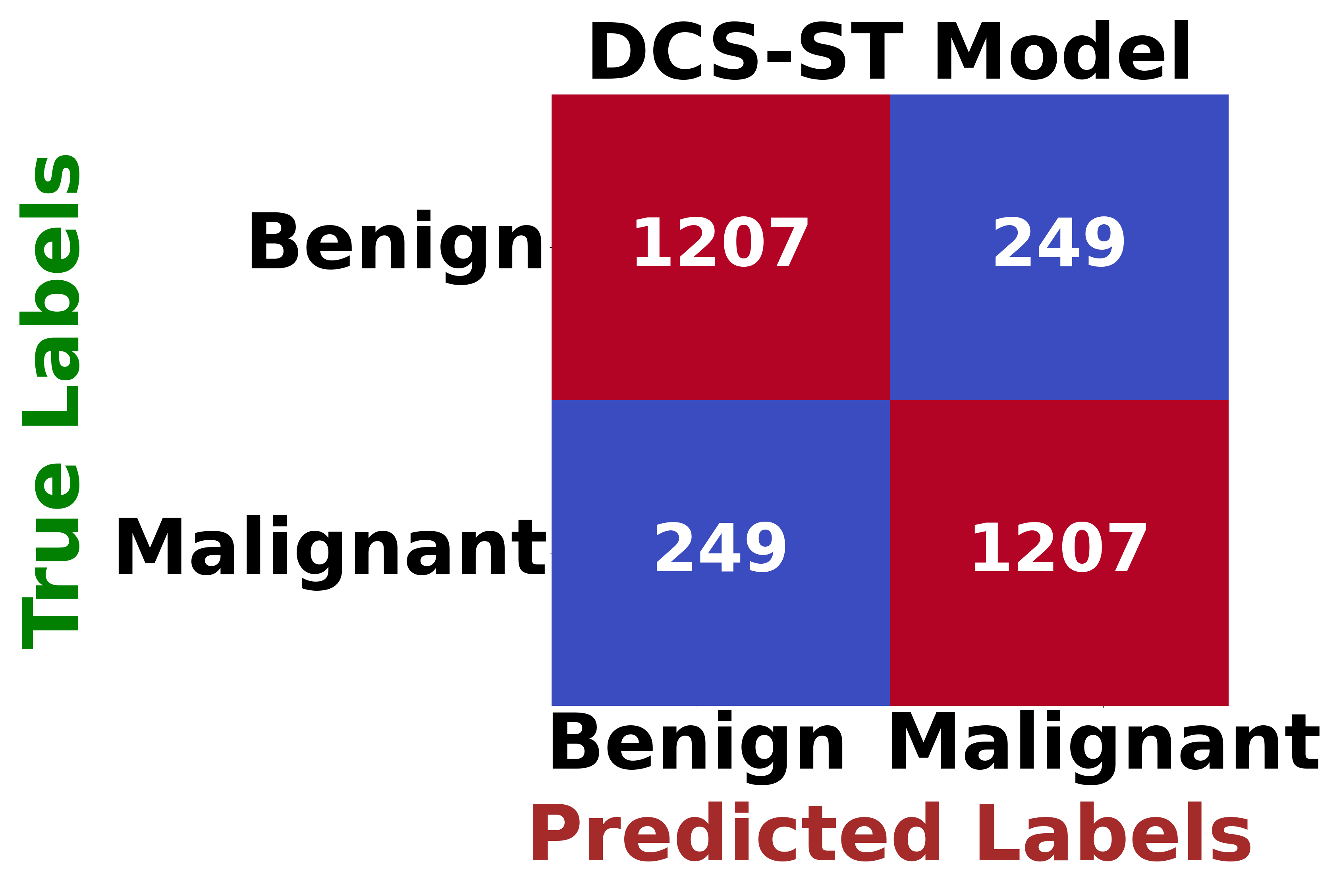}
    \end{minipage}%
    \hspace{0.02\textwidth}
    \begin{minipage}{0.22\textwidth}
        \centering
        \includegraphics[width=\linewidth]{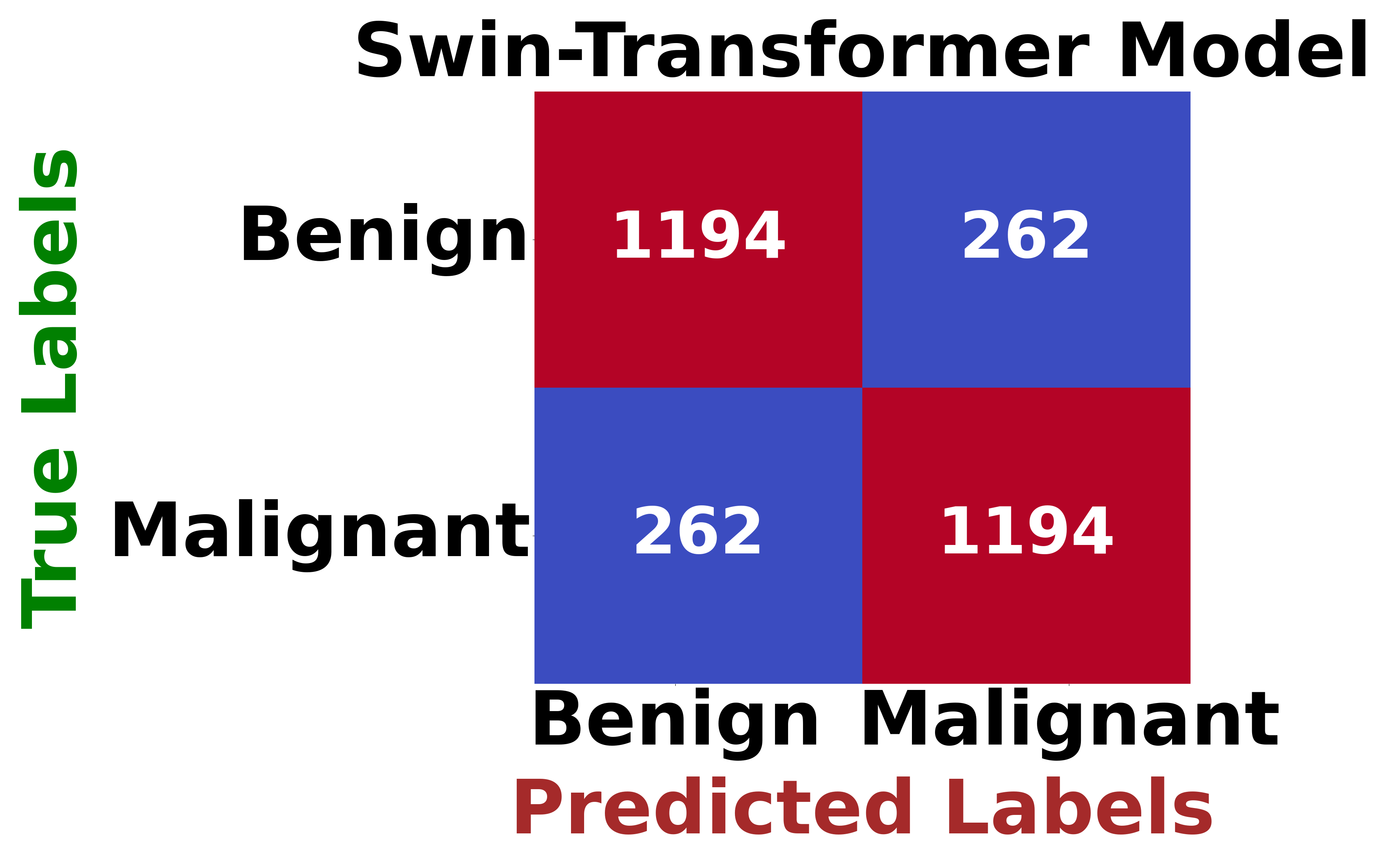}
    \end{minipage}

    \vspace{0.3cm} 
    \begin{minipage}{0.22\textwidth}
        \centering
        200x
    \end{minipage}%
    \hspace{0.29\textwidth} 
    \begin{minipage}{0.22\textwidth}
        \centering
         400x
    \end{minipage}
\caption{Comparison of the Confusion Matrices of the BreakHis Dataset}

\end{figure*}

\subsubsection{Performance Analysis}
On the BreakHis dataset at 40X magnification, DCS-ST achieves an AUC-ROC of $0.9187 \pm 0.0051$, a Balanced Accuracy of $0.8664 \pm 0.0053$, an F1 Score of $0.9004 \pm 0.0126$, and a Cohen’s Kappa of $0.6695 \pm 0.0244$, outperforming ViT ($0.8365 \pm 0.0159$) and SimCLR ($0.7956 \pm 0.0155$). At 400X, DCS-ST maintains superiority with an AUC-ROC of $0.9645 \pm 0.0243$. On Mini-DDSM, DCS-ST reports an AUC-ROC of $0.8334 \pm 0.0523$, surpassing Swin-UNETR ($0.8325 \pm 0.0293$), while on ICIAR2018, it achieves an AUC-ROC of $0.8781 \pm 0.0201$, outperforming ViT ($0.7584 \pm 0.0346$), demonstrating robust generalizability across tasks.

\subsection{Ablation Study}

```
To assess the effectiveness of our proposed DCS-ST model for breast cancer histopathology image classification with limited annotations, we conducted an ablation study by comparing it against a baseline Swin-Transformer\cite{b25} model, which leverages a hierarchical transformer architecture for multi-scale feature extraction. Experiments were performed on the BreakHis dataset (four-class classification at 40X, 100X, 200X, and 400X), the Mini-DDSM dataset (three-class classification), and the ICIAR2018 dataset (four-class classification), with results evaluated using AUC-ROC, Balanced Accuracy, F1 Score, and Cohen's Kappa, reported as mean $\pm$ standard deviation in Tables VI and VII.

Tables VI and VII demonstrate the consistent superiority of our DCS-ST model over the baseline Swin-Transformer across all evaluated datasets. On the BreakHis dataset at 200X magnification, DCS-ST achieves significantly enhanced performance with an AUC-ROC of $0.9528 \pm 0.0026$ and an F1 Score of $0.9339 \pm 0.0039$, compared to Swin-Transformer's $0.9430 \pm 0.0227$ and $0.9313 \pm 0.0026$, respectively. The performance gap widens further on the Mini-DDSM dataset, where DCS-ST attains an AUC-ROC of $0.8334 \pm 0.0523$, substantially outperforming Swin-Transformer's $0.7820 \pm 0.0173$. This pattern of superior performance continues with the ICIAR2018 dataset, where DCS-ST reaches an AUC-ROC of $0.8781 \pm 0.0201$ versus Swin-Transformer's $0.8279 \pm 0.0237$. The confusion matrices presented in Fig.5 and Fig.6 provide additional evidence of DCS-ST's enhanced discriminative capability, revealing higher true positive rates (exemplified by 1466 versus 1411 correct benign classifications at BreakHis 200X) and reduced misclassifications across all datasets. These comprehensive results underscore the effectiveness of our proposed dynamic window prediction and cross-scale attention mechanisms in capturing fine-grained histopathological features critical for accurate breast cancer classification.

\begin{figure*}[t]
    \centering
    
    \begin{minipage}{0.22\textwidth}
        \centering
        \includegraphics[width=\linewidth]{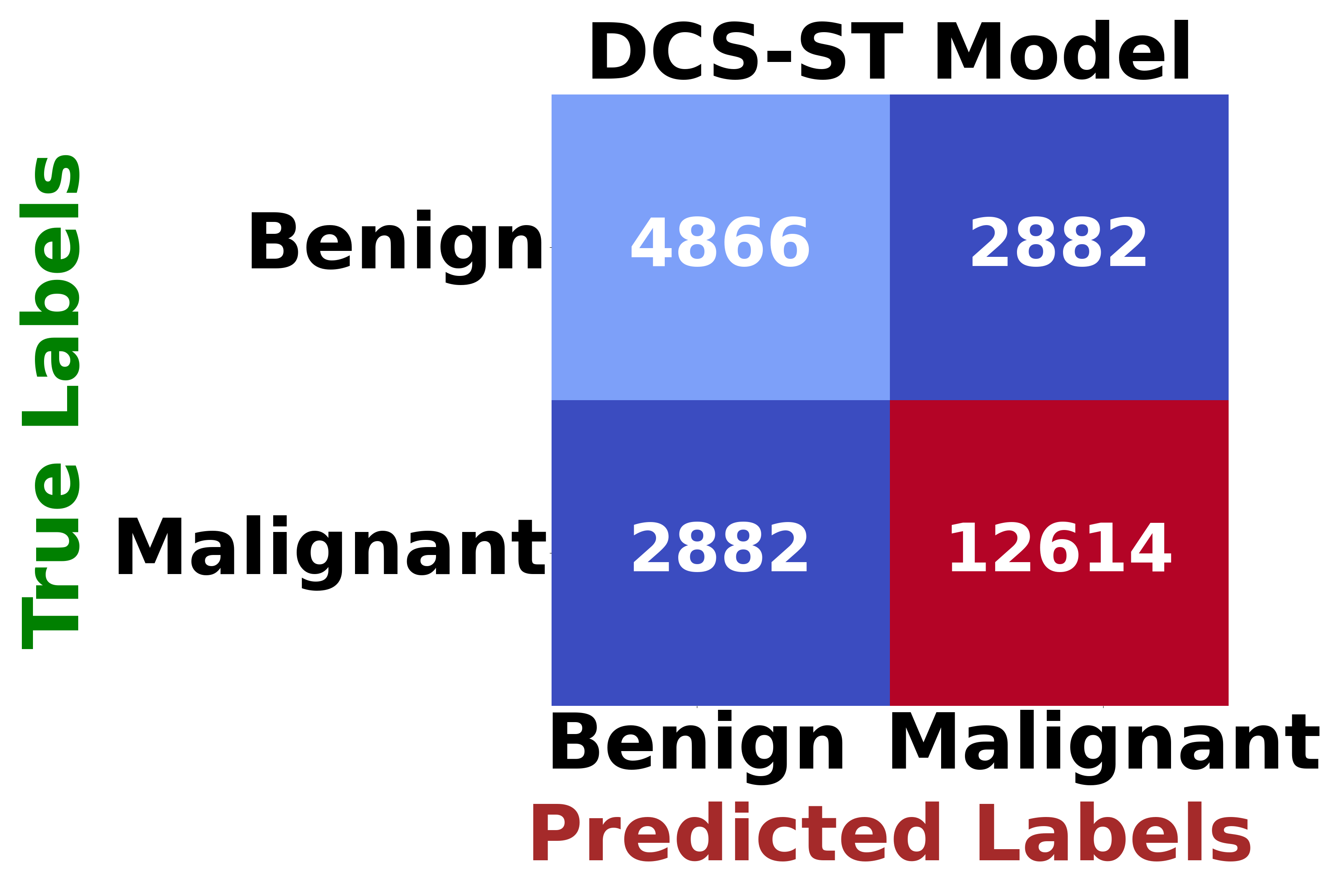}
    \end{minipage}%
    \hspace{0.02\textwidth}
    \begin{minipage}{0.22\textwidth}
        \centering
        \includegraphics[width=\linewidth]{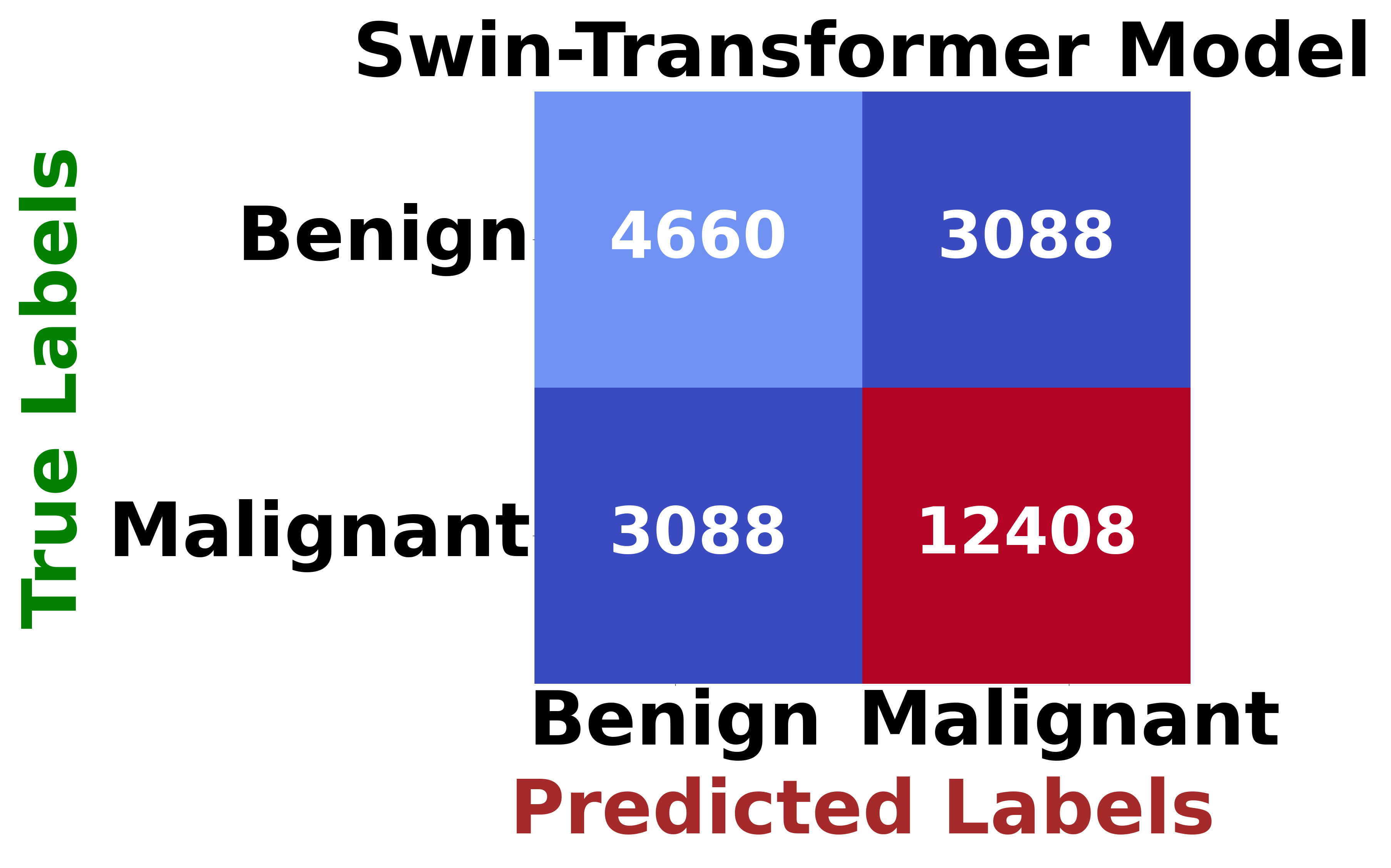}
    \end{minipage}%
    \hspace{0.05\textwidth}
    \begin{minipage}{0.22\textwidth}
        \centering
        \includegraphics[width=\linewidth]{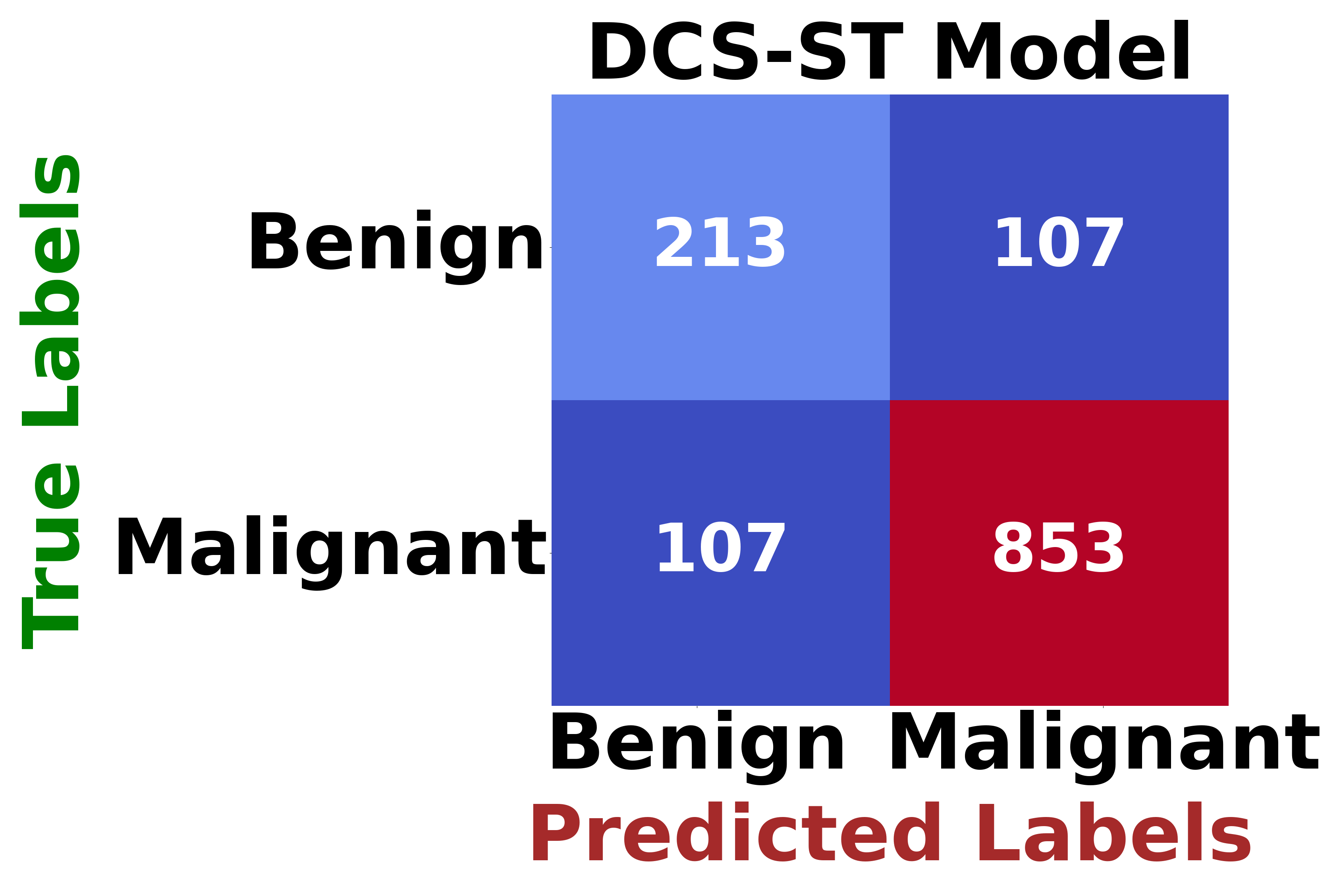}
    \end{minipage}%
    \hspace{0.02\textwidth}
    \begin{minipage}{0.22\textwidth}
        \centering
        \includegraphics[width=\linewidth]{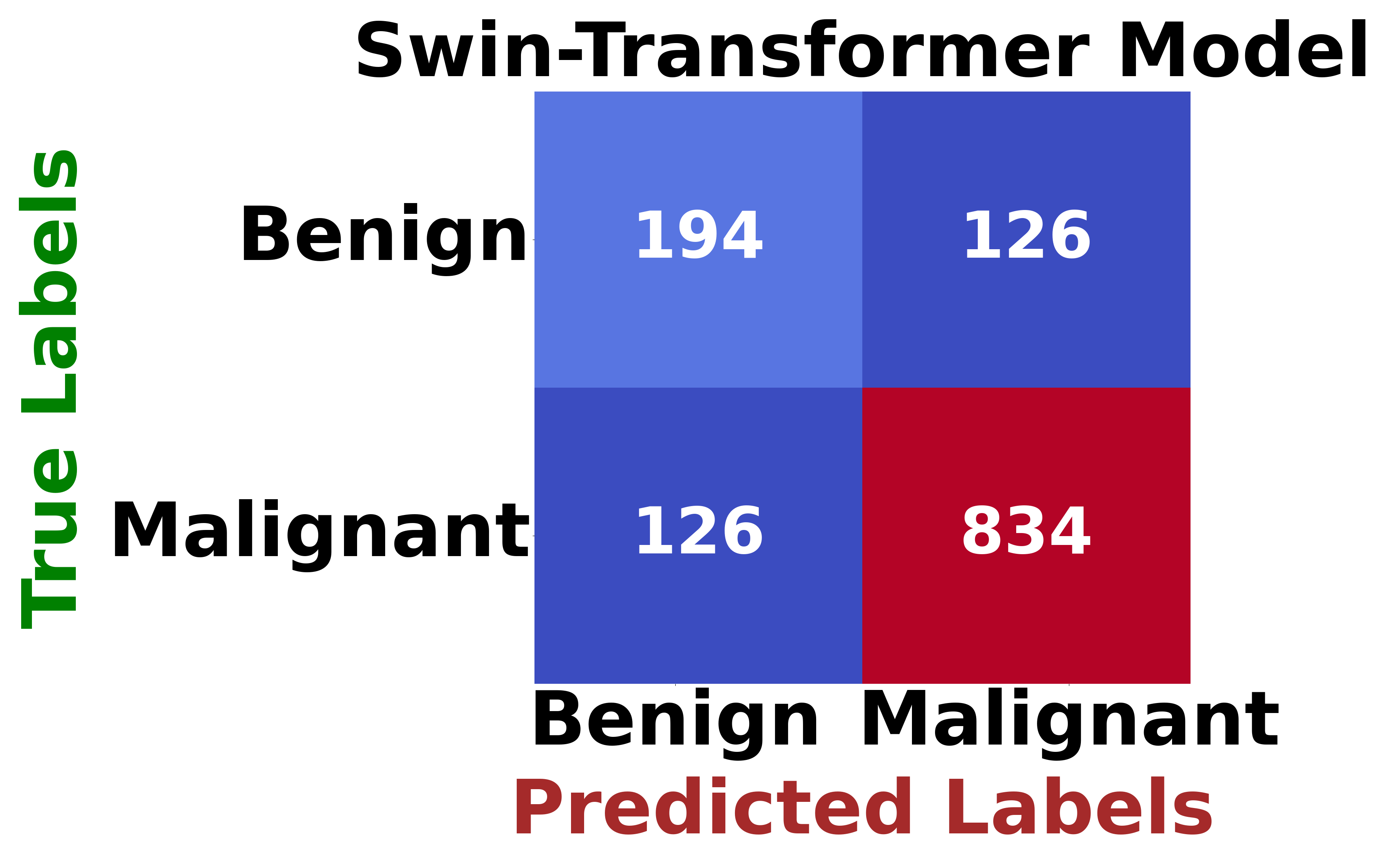}
    \end{minipage}

    \vspace{0.3cm} 
    \begin{minipage}{0.22\textwidth}
        \centering
         Mini-DDSM Dataset
    \end{minipage}%
    \hspace{0.29\textwidth} 
    \begin{minipage}{0.22\textwidth}
        \centering
         ICIAR2018 Dataset 
    \end{minipage}

\caption{Comparison of the Confusion Matrices of the Mini-DDSM Dataset (Three-Class Classification) and ICIAR2018 Dataset (Four-Class Classification)}
\end{figure*}

\subsection{Limitations}

While the DCS-ST model excels in classifying breast cancer histopathology images with limited annotations, it exhibits several limitations. The model's reliance on a pre-trained Swin-Transformer backbone and the computationally intensive cross-scale attention module may pose challenges for deployment on resource-constrained hardware, despite training on two NVIDIA T4 GPUs. Its semi-supervised learning strategy, which leverages 95\% unlabeled data, could be less effective if high-quality unlabeled data is scarce, as seen in the pseudo-label generation phase (confidence threshold of 0.9). Additionally, performance variations across magnification levels on the BreakHis dataset (e.g., lower Balanced Accuracy of $0.8253 \pm 0.0461$ at 400X) suggest potential struggles with unseen data variations. The model’s scalability to larger or more complex datasets, such as those beyond Mini-DDSM and ICIAR2018, remains uncertain, and its robustness against adversarial attacks is unverified. These limitations highlight the need for further research to enhance DCS-ST’s efficiency, generalizability, and resilience.

\section{Conclusion}

In this work, we introduce DCS-ST, a robust and efficient framework designed for classifying breast cancer histopathology images, even with limited annotation data. By incorporating a dynamic window predictor and a cross-scale attention module, DCS-ST achieves notable improvements in classification accuracy and computational efficiency. Experimental results across multiple datasets show that it outperforms existing methods. Beyond advancing medical image analysis, this framework also holds promising clinical potential as a supportive tool for early breast cancer detection.

\section*{References and Footnotes}

\section*{References}



\begin{thebibliography}{00}
\bibitem{b1} R. Krithiga, P. Geetha, ``Breast cancer detection, segmentation and classification on histopathology images analysis: a systematic review,'' in \emph{Archives of Computational Methods in Engineering}, vol. 28, no. 4, pp. 2607--2619, 2021.



\bibitem{b2} H. Jiang, Y. Yin, J. Zhang, W. Deng, C. Li, ``Deep learning for liver cancer histopathology image analysis: A comprehensive survey,'' in \emph{Engineering Applications of Artificial Intelligence}, vol. 133, 108436, 2024.


\bibitem{b3} H. Xu, Q. Xu, F. Cong, J. Kang, C. Han, Z. Liu, et al., ``Vision transformers for computational histopathology,'' in \emph{IEEE Reviews in Biomedical Engineering}, vol. 17, pp. 63--79, 2023.



\bibitem{b6}
F.~A. Spanhol, L.~S. Oliveira, C.~Petitjean, and L.~Heutte, ``A dataset for breast cancer histopathological image classification,'' \emph{Ieee transactions on biomedical engineering}, vol.~63, no.~7, pp. 1455--1462, 2015.

\bibitem{b7}
C.~D. Lekamlage, F.~Afzal, E.~Westerberg, and A.~Cheddad, ``Mini-DDSM: Mammography-based automatic age estimation,'' in \emph{Proc. 2020 3rd Int. Conf. Digital Medicine and Image Processing}, Nov. 2020, pp. 1--6.


\bibitem{b8}
G.~Aresta, T.~Araújo, S.~Kwok, S.~S. Chennamsetty, M.~Safwan, V.~Alex, B.~Marami, M.~Prastawa, M.~Chan, M.~Donovan, G.~Fernandez, J.~Zeineh, M.~Kohl, C.~Walz, F.~Ludwig, S.~Braunewell, M.~Baust, Q.~D. Vu, M.~N. To, E.~Kim, J.~T. Kwak, S.~Galal, V.~Sanchez-Freire, N.~Brancati, M.~Frucci, D.~Riccio, Y.~Wang, L.~Sun, K.~Ma, J.~Fang, I.~Kone, L.~Boulmane, A.~Campilho, C.~Eloy, and P.~Aguiar, ``BACH: Grand challenge on breast cancer histology images,'' \emph{Medical Image Analysis}, vol.~56, pp. 122--139, 2019.

\bibitem{b9}
Carrington, A. M., Manuel, D. G., Fieguth, P. W., Ramsay, T., Osmani, V., Wernly, B., ... \& Holzinger, A. (2022). Deep ROC analysis and AUC as balanced average accuracy, for improved classifier selection, audit and explanation. \textit{IEEE Transactions on Pattern Analysis and Machine Intelligence}, 45(1), 329-341.


\bibitem{b10}
Chicco, D., Tötsch, N., \& Jurman, G. (2021). The Matthews correlation coefficient (MCC) is more reliable than balanced accuracy, bookmaker informedness, and markedness in two-class confusion matrix evaluation. \textit{BioData Mining}, 14, 1-22.


\bibitem{b11}
Parekh, D. H., \& Dahiya, V. (2021). Predicting breast cancer using machine learning classifiers and enhancing the output by combining the predictions to generate optimal F1-score. \textit{Biomedical and Biotechnology Research Journal (BBRJ)}, 5(3), 331-334.



\bibitem{b12}
Wetstein, S. C., de Jong, V. M., Stathonikos, N., Opdam, M., Dackus, G. M., Pluim, J. P., ... \& Veta, M. (2022). Deep learning-based breast cancer grading and survival analysis on whole-slide histopathology images. \textit{Scientific Reports}, 12(1), 15102.



\bibitem{b13}
Zhang, L., Yang, L., Ma, T., Shen, F., Cai, Y., \& Zhou, C. (2021). A self-training semi-supervised machine learning method for predictive mapping of soil classes with limited sample data. \textit{Geoderma}, 384, 114809.

\bibitem{b14}
Nassar, I., Herath, S., Abbasnejad, E., Buntine, W., \& Haffari, G. (2021). All labels are not created equal: Enhancing semi-supervision via label grouping and co-training. In \textit{Proceedings of the IEEE/CVF Conference on Computer Vision and Pattern Recognition} (pp. 7241-7250).

\bibitem{b15}
Rong, S., Tu, B., Wang, Z., \& Li, J. (2023). Boundary-enhanced co-training for weakly supervised semantic segmentation. In \textit{Proceedings of the IEEE/CVF Conference on Computer Vision and Pattern Recognition} (pp. 19574-19584).
\bibitem{b16}
Sohn, K., Berthelot, D., Carlini, N., Zhang, Z., Zhang, H., Raffel, C. A., ... \& Li, C. L. (2020). Fixmatch: Simplifying semi-supervised learning with consistency and confidence. \textit{Advances in Neural Information Processing Systems}, 33, 596-608.

\bibitem{b17}
Peng, X., Peng, T., Yang, C., Ye, C., Chen, Z., \& Yang, C. (2024). Adversarial domain adaptation network with MixMatch for incipient fault diagnosis of PMSM under multiple working conditions. \textit{Knowledge-Based Systems}, 284, 111331.

\bibitem{b18}
Pham, D. H., Nguyen, A. D., \& Nguyen, H. N. (2024). GAN-based data augmentation and pseudo-label refinement with holistic features for unsupervised domain adaptation person re-identification. \textit{Knowledge-Based Systems}, 288, 111471.

\bibitem{b19}
Abdulrazzaq, M. M., Ramaha, N. T., Hameed, A. A., Salman, M., Yon, D. K., Fitriyani, N. L., ... \& Lee, S. W. (2024). Consequential advancements of self-supervised learning (SSL) in deep learning contexts. \textit{Mathematics}, 12(5), 758.
\bibitem{b20}
Pani, K., \& Chawla, I. (2024). A hybrid approach for multi modal brain tumor segmentation using two phase transfer learning, SSL and a hybrid 3DUNET. \textit{Computers and Electrical Engineering}, 118, 109418.
\bibitem{b41}
Cao, H., Wang, Y., Chen, J., Jiang, D., Zhang, X., Tian, Q., \& Wang, M. (2022, October). Swin-unet: Unet-like pure transformer for medical image segmentation. In \textit{European Conference on Computer Vision} (pp. 205-218). Cham: Springer Nature Switzerland.
\bibitem{b42}
Nayak, D. R. (2024). RDTNet: A residual deformable attention based transformer network for breast cancer classification. \textit{Expert Systems with Applications}, 249, 123569.


\bibitem{b43}
Yang, B., Li, J., Wong, D. F., Chao, L. S., Wang, X., \& Tu, Z. (2019, July). Context-aware self-attention networks. In \textit{Proceedings of the AAAI Conference on Artificial Intelligence} (Vol. 33, No. 01, pp. 387-394).
\bibitem{b44}
Chen, P. H., Hsieh, J. W., Hsieh, Y. K., Chang, C. W., \& Huang, D. Y. (2025). Cross-Scale Overlapping Patch-Based Attention Network for Road Crack Detection. \textit{IEEE Transactions on Intelligent Transportation Systems}.
\bibitem{b45}
Shen, T., Li, H., \& Huang, X. (2010). Active volume models for medical image segmentation. \textit{IEEE Transactions on Medical Imaging}, 30(3), 774-791.
\bibitem{b46}
Zhao, Q., Wu, D., \& Tian, J. (2024, November). PCASNet: Polarized Cross-scale Attention Self-attention Network for Lightweight Ultrasound Medical Image Segmentation. In \textit{2024 International Conference on Image Processing, Computer Vision and Machine Learning (ICICML)} (pp. 326-330). IEEE.









\bibitem{b21} T. Chen, S. Kornblith, M. Norouzi, and G. Hinton, ``A simple framework for contrastive learning of visual representations,'' in \emph{International Conference on Machine Learning}, vol. 119, pp. 1597--1607, Nov. 2020.


\bibitem{b22} K. Sohn, D. Berthelot, N. Carlini, Z. Zhang, H. Zhang, C. A. Raffel, ... \& C. L. Li, "Fixmatch: Simplifying semi-supervised learning with consistency and confidence," in \emph{Advances in Neural Information Processing Systems}, vol. 33, pp. 596--608, Nov. 2020.

\bibitem{b23} H. Feng, Y. Jia, R. Xu, M. Prasad, A. Anaissi, \& A. Braytee, "Integration of self-supervised BYOL in semi-supervised medical image recognition," in \emph{International Conference on Computational Science}, pp. 163--170, Cham: Springer Nature Switzerland, Jun. 2024.



\bibitem{b24} A. Dosovitskiy, "An image is worth 16x16 words: Transformers for image recognition at scale," arXiv preprint arXiv:2010.11929, 2020.



\bibitem{b25} Z. Liu, Y. Lin, Y. Cao, H. Hu, Y. Wei, Z. Zhang, et al., "Swin transformer: Hierarchical vision transformer using shifted windows," in Proceedings of the IEEE/CVF International Conference on Computer Vision, 2021, pp. 10012-10022.




\bibitem{b26} Z. Dai, H. Liu, Q. V. Le, and M. Tan, ``CoAtNet: Marrying Convolution and Attention for All Data Sizes,'' in \emph{Advances in Neural Information Processing Systems}, vol. 34, pp. 3965--3977, 2022.


\bibitem{b27} J. Kim, H. S. Lee, I. S. Song, and K. H. Jung, ``DeNTNet: Deep Neural Transfer Network for the detection of periodontal bone loss using panoramic dental radiographs,'' in \emph{Scientific Reports}, vol. 9, no. 1, p. 17615, 2019.



\bibitem{b28} O. N. Manzari, H. Ahmadabadi, H. Kashiani, S. B. Shokouhi, and A. Ayatollahi, ``MedViT: a robust vision transformer for generalized medical image classification,'' in \emph{Computers in Biology and Medicine}, vol. 157, p. 106791, 2023.


\bibitem{b29} M. Tan and Q. Le, ``Efficientnetv2: Smaller models and faster training,'' in \emph{International Conference on Machine Learning}, pp. 10096--10106, Jul. 2021.

\bibitem{b30} A. Hatamizadeh, V. Nath, Y. Tang, D. Yang, H. R. Roth, and D. Xu, ``Swin UNETR: Swin transformers for semantic segmentation of brain tumors in MRI images,'' in \emph{International MICCAI Brainlesion Workshop}, pp. 272--284, Sep. 2021.



\bibitem{b31} W. L. Ding, X. J. Zhu, K. Zheng, J. L. Liu, and Q. H. You, ``A multi-level feature-fusion-based approach to breast histopathological image classification,'' in \emph{Biomedical Physics \& Engineering Express}, vol. 8, no. 5, p. 055002, 2022.





\end{thebibliography}
\end{document}